\documentclass[conference]{IEEEtran}
\usepackage{cite}
\usepackage{amsmath,amssymb,amsfonts}
\usepackage{algorithmic}
\usepackage{graphicx}
\usepackage{subcaption}
\usepackage{textcomp}
\usepackage[table,xcolor]{xcolor}

\usepackage{url}
\usepackage{diagbox}
\usepackage{float}
\usepackage[numbers]{natbib}
\usepackage{cleveref}
\usepackage{arydshln}
\usepackage{makecell}
\usepackage{verbatim}
\usepackage{placeins}

\def\BibTeX{{\rm B\kern-.05em{\sc i\kern-.025em b}\kern-.08em
    T\kern-.1667em\lower.7ex\hbox{E}\kern-.125emX}}
\begin{document}
\title{Reducing Texture Bias of Deep Neural Networks via Edge Enhancing Diffusion}

\author{\IEEEauthorblockN{1\textsuperscript{st} Edgar Heinert}
\IEEEauthorblockA{\textit{School of Mathematics and Natural Sciences} \\
\textit{University of Wuppertal}\\
Wuppertal, Germany \\
heinert@uni-wuppertal.de} \\
\IEEEauthorblockN{3\textsuperscript{rd} Kira Maag}
\IEEEauthorblockA{\textit{Faculty of Mathematics and Natural Sciences} \\
\textit{Technical University of Berlin}\\
Berlin, Germany \\
maag@tu-berlin.de}
\and
\IEEEauthorblockN{2\textsuperscript{nd} Matthias Rottmann}
\IEEEauthorblockA{\textit{School of Mathematics and Natural Sciences} \\
\textit{University of Wuppertal}\\
Wuppertal, Germany \\
rottmann@uni-wuppertal.de} \\
\IEEEauthorblockN{4\textsuperscript{th} Karsten Kahl}
\IEEEauthorblockA{\textit{School of Mathematics and Natural Sciences} \\
\textit{University of Wuppertal}\\
Wuppertal, Germany \\
kkahl@uni-wuppertal.de}
}
\maketitle
\begin{abstract}
Convolutional neural networks (CNNs) for image processing tend to focus on localized texture patterns, commonly referred to as texture bias. While most of the previous works in the literature focus on the task of image classification, we go beyond this and study the texture bias of CNNs in semantic segmentation. In this work, we propose to train CNNs on pre-processed images with less texture to reduce the texture bias. Therein, the challenge is to suppress image texture while preserving shape information. To this end, we utilize edge enhancing diffusion (EED), an anisotropic image diffusion method initially introduced for image compression, to create texture reduced duplicates of existing datasets. 
Extensive numerical studies are performed with both CNNs and vision transformer models trained on original data and EED-processed data from the Cityscapes dataset and the CARLA driving simulator. We observe strong texture-dependence of CNNs and moderate texture-dependence of transformers. Training CNNs on EED-processed images enables the models to become completely ignorant with respect to texture, demonstrating resilience with respect to texture re-introduction to any degree. Additionally we analyze the performance reduction in depth on a level of connected components in the semantic segmentation and study the influence of EED pre-processing on domain generalization as well as adversarial robustness.
\end{abstract}


\begingroup
\renewcommand\thefootnote{}
\footnotetext{This work has been supported by the German Federal Ministry of Education and Research within the junior research group project ``UnrEAL'', grant no.\ 01IS22069 as well as by the German Federal Ministry for Economic Affairs and Climate Action within the project ``KI Delta Learning - Scalable AI for Automated Driving'', grant no.\ 19A19013Q.}
\endgroup

\begin{figure}
    \centering
    \includegraphics[width=0.45\textwidth]{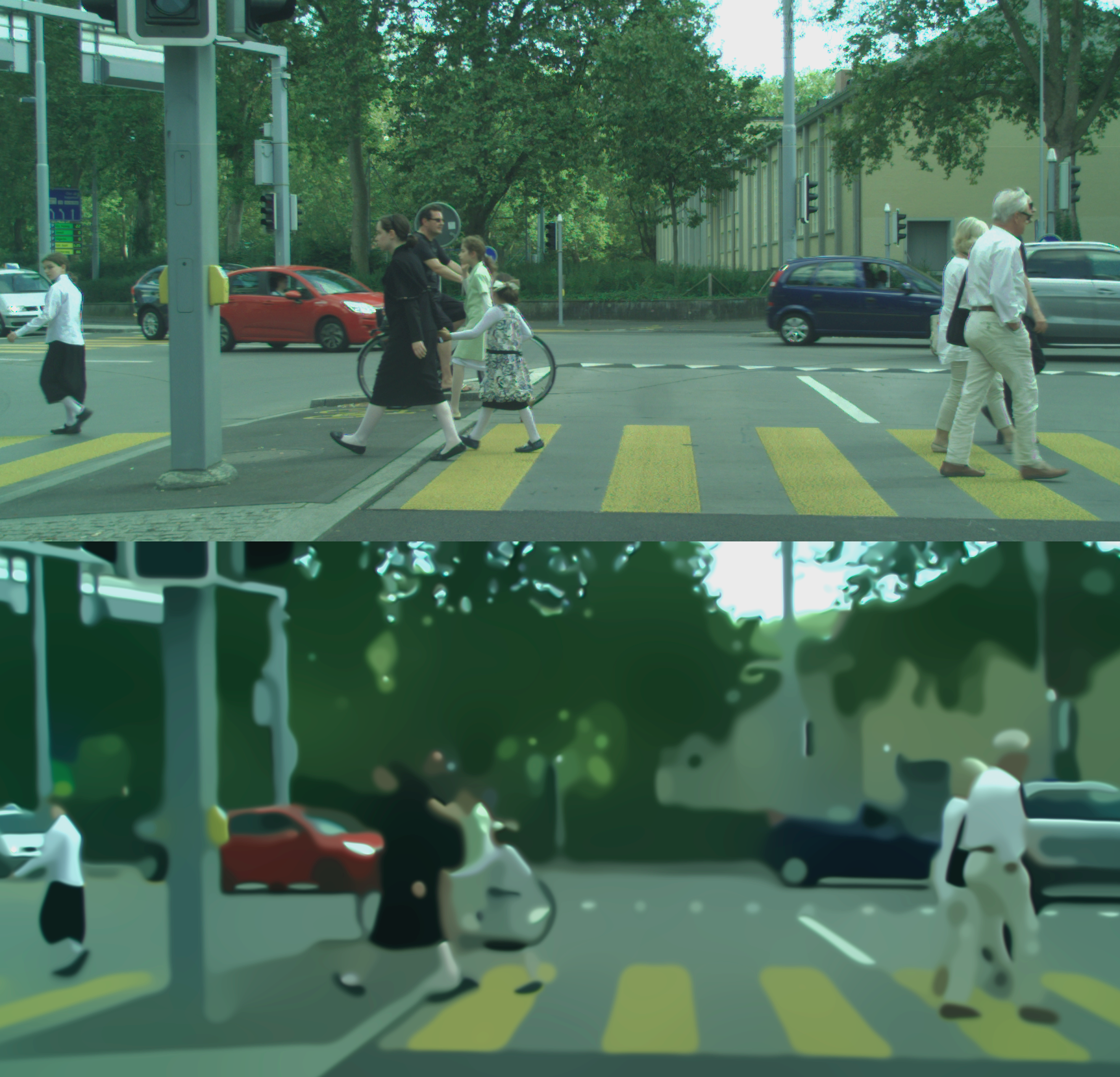}
    \caption{A Cityscapes image (top) and its EED-processed counterpart (bottom). Texture is removed to a great extent by EED while shapes and semantic meaning are preserved.}
    \label{fig:city_diff}
    \vspace{-10pt}
\end{figure}

\section{Introduction}
Convolutional neural networks (CNNs, \citep{lecun1998gradient,krizhevsky2012imagenet}) are a class of well established architectures in the field of deep learning and have become a cornerstone in various computer vision tasks, such as image classification and semantic segmentation. However, despite their remarkable success, CNNs are prone to strong biases. One prevalent bias observed in these networks is the texture bias \citep{b9,b10,b11}. Texture bias refers to a tendency to rely on local textural patterns rather than shape or structural information when making predictions or classifications. This bias often leads CNNs to prioritize texture-related details while potentially overlooking higher-level shape information or features. In contrast to CNNs, human visual perception is more reliant on shape information rather than texture \cite{jacob2021qualitative}.

A more recent breakthrough is the application of transformer architectures to computer vision \cite{dosovitskiy2020image}. Transformers have initially been designed for natural language processing tasks \cite{vaswani2017attention} where local proximity of words does not guarantee semantic connection and distance does not necessarily mean unrelatedness. While CNNs process images using filters that capture local patterns, building up a hierarchical representation of features, transformers utilize self-attention mechanisms to directly model relationships between all parts of the image. Thanks to this architectural difference, transformers are said to be less texture biased than CNNs and to be more reliant on shape information. Understanding and altering shape and texture biases in both CNNs and transformers have become key areas of interest in current computer vision research.

\begin{figure*}
    \centering
    \begin{minipage}[b]{0.32\textwidth}
        \includegraphics[trim={14cm 10cm 22cm 0cm},clip,width=\linewidth]{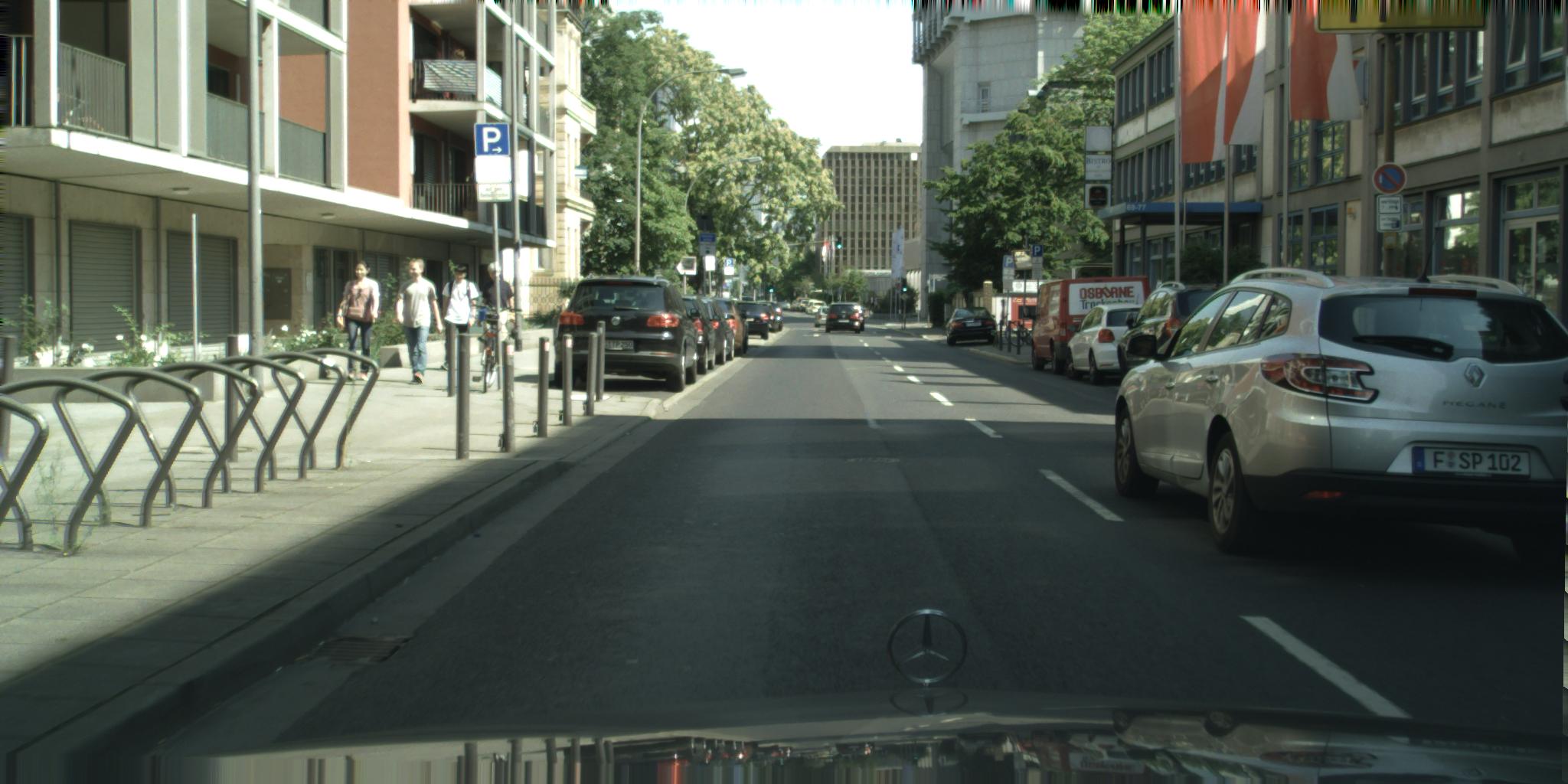}
    \end{minipage}
    \hfill 
    \begin{minipage}[b]{0.32\textwidth}
        \includegraphics[trim={14cm 10cm 22cm 0cm},clip,width=\linewidth]{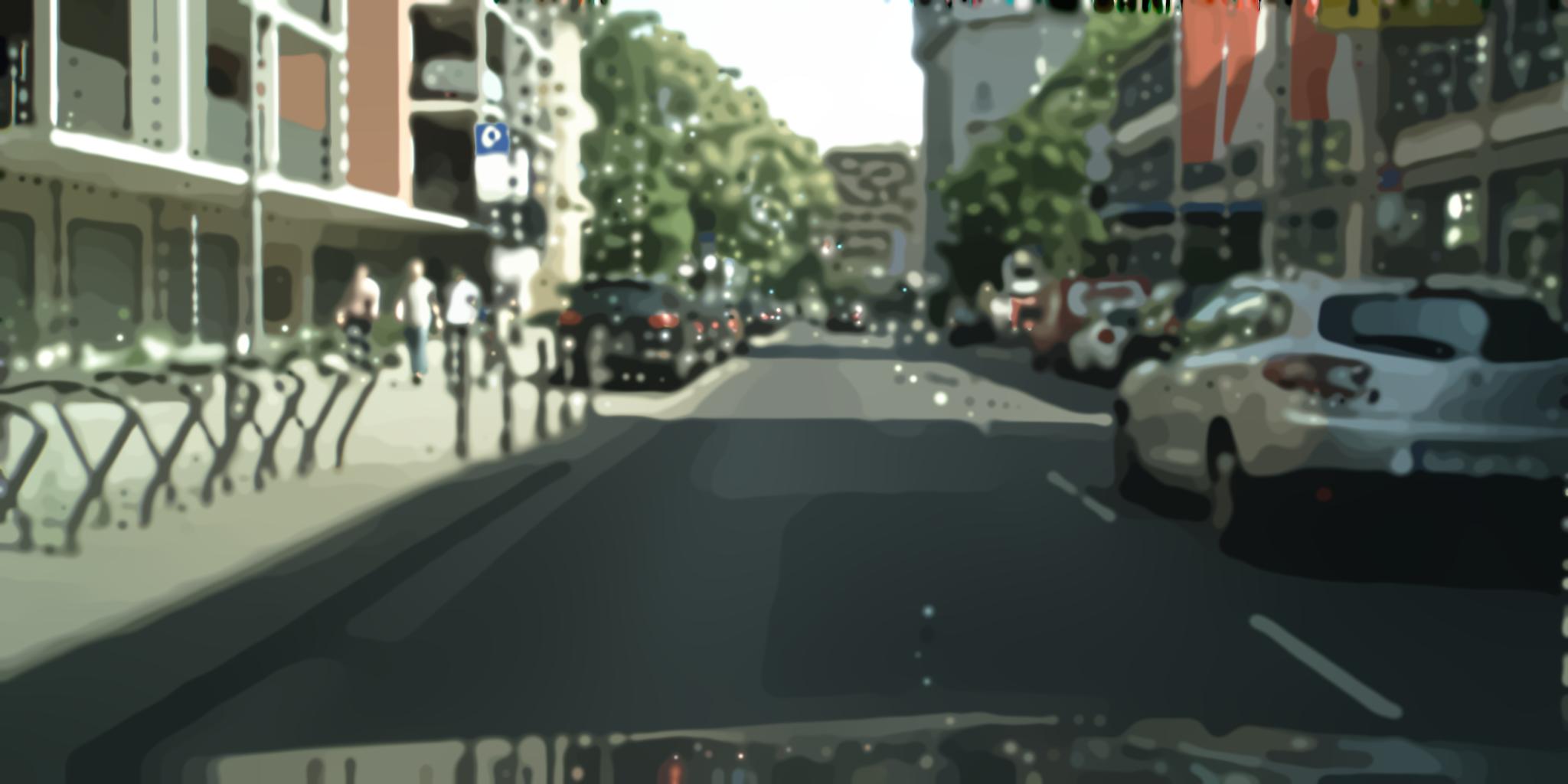}
    \end{minipage}
    \hfill
    \begin{minipage}[b]{0.32\textwidth}
        \includegraphics[trim={14cm 10cm 22cm 0cm},clip,width=\linewidth]{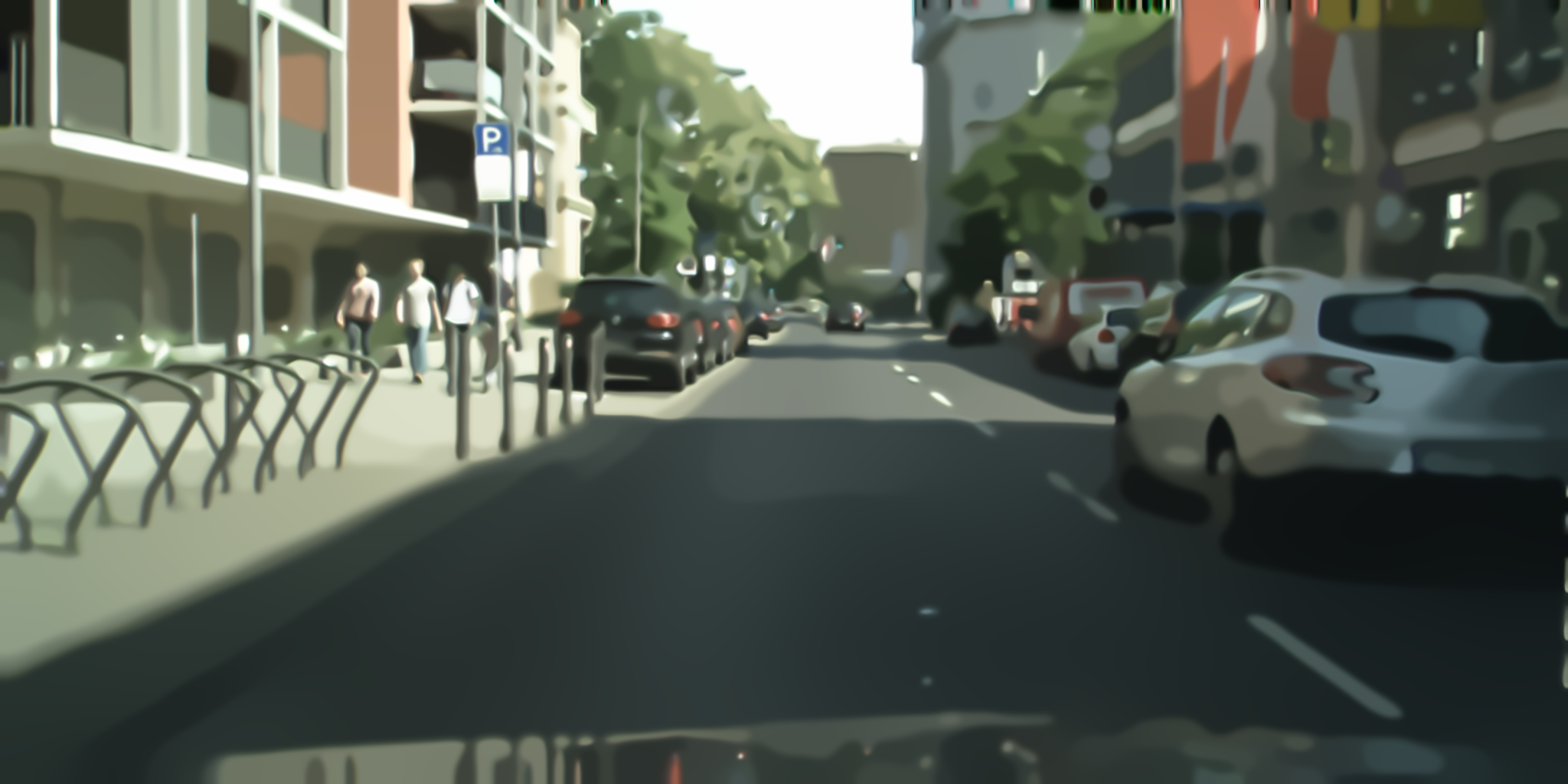}
    \end{minipage}
    \caption{Visual Comparison of an original Cityscapes image (left), EED  without orientation smoothing (mid) and with orientation smoothing (right). Orientation smoothing preserves shapes while preventing circular singularities.}
    \label{fig:orient_smoothing}
\end{figure*}

In this work, we propose a form of anisotropic diffusion, namely edge enhancing diffusion (EED, \cite{b1,b3,b4}; cf.\  \cref{fig:city_diff}), introduced initially in the field of camera-based computer vision. To avoid image artifacts, we extend EED by a stabilizing orientation smoothing, which was initially used in the context of coherence enhancing diffusion \cite{weickert1999coherence}. Thereby, to the best of our knowledge, we provide a new EED method for anisotropic image diffusion that effectively reduces texture while maintaining shape information in images. Cf.\ \cref{fig:orient_smoothing} for a visual example. We utilize EED to study and reduce the texture bias of deep neural networks (DNNs), primarily in the context of semantic segmentation. 

EED is a partial differential equation (PDE) based diffusion method that stems from classical signal theory and has originally been used in the context of image compression for the reconstruction of an image from only a few selected pixels. The used diffusion kernel is a variation of the Laplace operator and allows color information to diffuse along edges while preventing diffusion across them. In a recent publication it has been shown that a related and lightly applied form of anisotropic diffusion, used as a data augmentation technique, can help mitigate the domain gap from real to synthetic data, i.e., the domain gap between ImageNet to SketchImageNet \cite{b8}.

In our experiments, we study the texture bias of semantic segmentation networks, CNNs as well as transformers, on semantic segmentation datasets. To this end, we create EED-processed duplicates of Cityscapes \cite{b18} (showing German urban street scenes) and data obtained from the CARLA driving simulator \cite{b19}. Besides an examination of texture bias, we demonstrate the usefulness of the diffused images in gaining texture robustness for both CNN and transformer architectures. We show that training such DNNs on EED pre-processed images results in networks that are robust w.r.t.\ the reintroduction of said texture. In contrast to that, we observe that DNNs trained on original (not diffused) images are very sensitive to the removal of texture. Due to the lack of texture information in training images, the networks trained on EED data are less texture biased than networks that have been trained on the original data. We complement our semantic segmentation experiments with a few classification experiments on a dataset derived from Cityscapes to demonstrate the generality of EED for texture bias analysis and reduction.

Our contribution can be summarized as follows:
\begin{itemize}
    \item For the first time, we introduce an EED-based pre-processing with orientation smoothing for deep learning to analyze and reduce texture bias in image classification and semantic segmentation. A careful parameter selection provides EED duplicates of image classification and semantic segmentation datasets with reduced texture while preserving shape.
    \item Using EED, we report strong texture bias of CNNs in semantic segmentation and image classification and, on the other hand, confirm a rather moderate texture bias for transformers in semantic segmentation.
    \item In both cases, for image classification and semantic segmentation, we demonstrate that EED pre-processing makes DNNs almost ignorant w.r.t.\ local texture patterns, while the task performance loss remains moderate. A detailed segment-level analysis reveals that much of the task performance loss can be attributed to over-diffusing in visually challenging situations.
\end{itemize}

We make our code, including an efficient GPU-capable torch implementation of EED, publicly available on GitHub under \url{https://github.com/eheinert/reducing-texture-bias-of-dnns-via-eed/}.

\section{Related Work}
In this section, we first provide a brief overview over the evolution of anisotropic and edge enhancing diffusion and in addition describe how our contribution fits into the existing body of work regarding DNN texture and shape biases.

\textbf{Anisotropic diffusion.} Anisotropic diffusion is a PDE-based, classical signal theory image diffusion technique that has been first introduced by Perona and Malik in 1994 \cite{b1} and has been used for noise reduction, edge detection and non-AI image segmentation in~\cite{b5,b6}. Weickert suggested further developments in~\citep{b3,b4} and proposed the special cases of edge enhancing diffusion (EED) and coherence enhancing diffusion \cite{b2}. EED has been shown to be effective in lossy image compression in~\cite{b7}. Regarding application in the context of AI, it has recently been shown in~\cite{b8} that anisotropic diffusion, as proposed by Perona and Malik, when used as a data augmentation technique can help to mitigate the domain gap from real-world images to drawn images in image classification using the data sets ImageNet and SketchImageNet. In contrast to this first appearance of anisotropic diffusion in AI, our work focuses on in-domain texture robustness rather than closing domain gaps to the artificial domain. We employ fully diffused datasets as training data instead of a data augmentation approach and test the application on the task of semantic segmentation of street scenes.

\textbf{Biases of DNNs.} First seminal work on biases of CNNs was conducted as early as 2014 when Zeiler and Fergus used deconvolutional layers to visualize the features learned by CNNs, which revealed a mixture of texture and shape understanding \cite{b9}. This was followed by a first quantitative analysis of texture bias in~\cite{b10}. In 2018, A landmark paper by Geirhos et al.\ demonstrated that CNNs trained on standard datasets like ImageNet are biased towards recognizing textures rather than shapes. They used stylized images to show that CNNs often prioritize texture information over shape \cite{b11}. Transformer architectures on the other hand show a tendency towards a stronger shape bias than CNNs \cite{Zhang2021Delving} and the errors of attention-based architectures seem to be more consistent with those of humans \cite{tuli2021convolutional}.

Apart from the mentioned anisotropic diffusion augmentation, there is a whole variety of approaches to reducing the texture bias in CNNs that include style domain adversarial training \cite{b12} and a contrastive learning approach\cite{b13} where images with the same texture but diminished semantics are used as negatives. Furthermore, several other data augmentation techniques such as style transfer \cite{b14}, edge deformation \cite{b16} and shape focused augmentation \cite{b15}, have demonstrated to reduce texture biases. Of those works, only the style transfer augmentation \cite{b14} works with street scenes, although in the context of 2D object detection and with a focus on domain shifts. The texture reduction technique meanshift \cite{b17} was used in order to show that the bias developed by CNNs is task dependent.

We are the first to use EED as a data pre-processing technique in order to reduce texture bias in DNNs and distinguish our work by the focus on texture robustness, the downstream task of semantic segmentation and the application on street scene data as considered only in \cite{b14}.

\section{Edge Enhancing Diffusion with Smoothed Orientation}

At the core of the data pre-processing used in our work is the PDE-based EED method proposed by Weickert et al.\ \cite{b2}. The idea is to make color values spatially diffuse parallel to edges while preventing diffusion perpendicular to edges. For the sake of simplicity, we review the construction of EED for gray scale images. For a given gray scale image, we assume that a given image in matrixform is the discretization of a continuous function $f$ on a rectangle $\Omega := [0,n]\times [0,m]$. The diffusion takes place by solving the following PDE with respect to time, using $f$ as its initial value and zero valued Neumann boundary conditions:
\begin{align}
\delta_t u &:= \nabla^T g(\nabla u_\sigma \nabla u_\sigma^T)\nabla u, \\ u(x,0)&=f(x) \, . \nonumber
\end{align}
Here $u_\sigma:=  K_\sigma * u$ is the image smoothed with a Gaussian kernel with standard deviation $\sigma$. Note that for $g(\nabla u_\sigma \nabla u_\sigma^T) = I $ this would be the Laplace operator and the diffusion indeed isotropic. The matrix function $g$ is the Charbonnier diffusivity function \cite{charbonnier1997deterministic}
\begin{align}
g(s):=\frac{1}{\sqrt{1+\frac{s}{\kappa^2}}}
\end{align}
for some contrast parameter $\kappa>0$ and $\nabla$ consists of the partial derivatives w.r.t.\ the physical variables, not the time parameter. The diffusion tensor $ g(\nabla u_\sigma \nabla u_\sigma^T)$ is a $2 \times 2$ matrix with eigenvectors parallel and orthogonal to the gradient $\nabla u_\sigma$. The corresponding eigenvalues are $g(|\nabla u_\sigma|^2)$ and $1$ as its argument has the eigendecomposition
\begin{align}
\biggl\lbrack \frac{\nabla u_\sigma}{||\nabla u_\sigma||}, \frac{\nabla u_\sigma^\bot}{||\nabla u_\sigma||} \biggr\rbrack 
\left[ \begin{array}{rr}
||\nabla u_\sigma||^2 & 0  \\ 
0 \hspace*{6mm} & 0  \\ 
\end{array}\right]
\biggl\lbrack \frac{\nabla u_\sigma}{||\nabla u_\sigma||}, \frac{\nabla u_\sigma^\bot}{||\nabla u_\sigma||} \biggr\rbrack^T.
\end{align}
Therefore, more information is diffused orthogonally to the gradient, i.e., along edges, and only relatively little diffusion happens across edges, depending on how small $\kappa$ is chosen. Numerically, the diffusion is performed via consecutive gradient descent steps on a discretized version of the energy
\begin{align} \label{eq:eed}
E(u)=\frac{1}{2} \int_\Omega \nabla u^T g(\nabla u_\sigma \nabla u_\sigma^T) \, \nabla  u \, \textrm{d}x \,\textrm{d}y.
\end{align}

For the discretization we use a non-standard finite differences approach as discussed by Weickert et al.\ \cite[p. 380-391]{weickert20132}. An extension for multiple channel images can be found in \cite[p. 321]{weickert2006tensor}. In order to avoid circular singularities in the diffused images we apply an additional element-wise Gaussian smoothing to the $2 \times 2$ orientations $\nabla u_\sigma \nabla u_\sigma^T$ in the form of $K_\sigma*(\nabla u_\sigma \nabla u_\sigma^T)$ as proposed for coherence enhancing diffusion \cite{weickert1999coherence}. That is, we perform a gradient descent iteration to minimize the energy
\begin{align} \label{eq:eedgs}
\tilde{E}(u)=\frac{1}{2} \int_\Omega \nabla u^T g\left( K_\sigma * (\nabla u_\sigma \nabla u_\sigma^T)  \right) \, \nabla  u \, \textrm{d}x \,\textrm{d}y.
\end{align}
We provide a visual comparison of EED \eqref{eq:eed} and EED with orientation smoothing \eqref{eq:eedgs} in \cref{fig:orient_smoothing}, demonstrating how the latter contributes to the reduction of image artifacts in terms of circular singularities.

Throughout our experiments, we use EED as a pre-processing method to create texture reduced duplicates of images and datasets. We only process the camera images of a given dataset while the corresponding labels remain the original ones. Details are provided in \cref{subsec:datasets}.

\section{Experiments}
In this section we provide numerical experiments showing the effect of texture bias reduction by training DNNs on EED-processed data.
Firstly, we introduce the datasets we use for our experiments and elaborate on the application of EED to generate various duplicates with reduced texture.
Thereafter, we showcase the generality of our approach by presenting results for the task of image classification. Subsequently, we focus on our main subject, reducing texture bias in semantic segmentation DNNs. The corresponding experiments include comparative evaluations of networks trained on both original and EED-processed data using the Deeplabv3+ \cite{b20} CNN and Segformer b1 \cite{xie2021segformer} transformer architectures, an ablation study on EED time steps $t$, and a segment-wise analysis to assess the performance of a Deeplabv3+ trained on EED data and evaluated on both original and EED datasets in greater detail. Finally, we study the adversarial robustness of semantic segmentation DNNs trained on EED-processed data and compare it with DNNs trained on original data. Note that all networks have been trained from scratch, i.e., we completely omitted ImageNet pre-training to provide unbiased results.

\subsection{Datasets}\label{subsec:datasets}

Throughout our experiments we use data from the Cityscapes dataset \cite{b18} containing street scene images of German urban environments, labeled with semantic segmentation masks. In addition we use 4000 images obtained from the CARLA driving simulator \cite{b19}, for which semantic segmentation masks are available as well. More precisely, we use the 2975 finely labeled training images of Cityscapes for training and the 500 finely labeled validation images for testing / evaluation. The 4000 images with labels from the CARLA simulator were recorded from the towns 01, 02, 03, 04, 05 and 07. Towns 02, 03, 04, 07 have been used for training and town 01 and 05 for evaluation. This results in 3000 training and 1000 test images. In order to make the results from Cityscapes and CARLA comparable, we did not use the full number of classes available in Cityscapes and CARLA. Instead, we used 14 common classes of the two datasets. The remaining classes can be found in \cref{tab:classwise}.

\begin{figure*}
    \centering
    \begin{subfigure}[t]{0.33\textwidth}
        \centering
        \includegraphics[width=\linewidth]{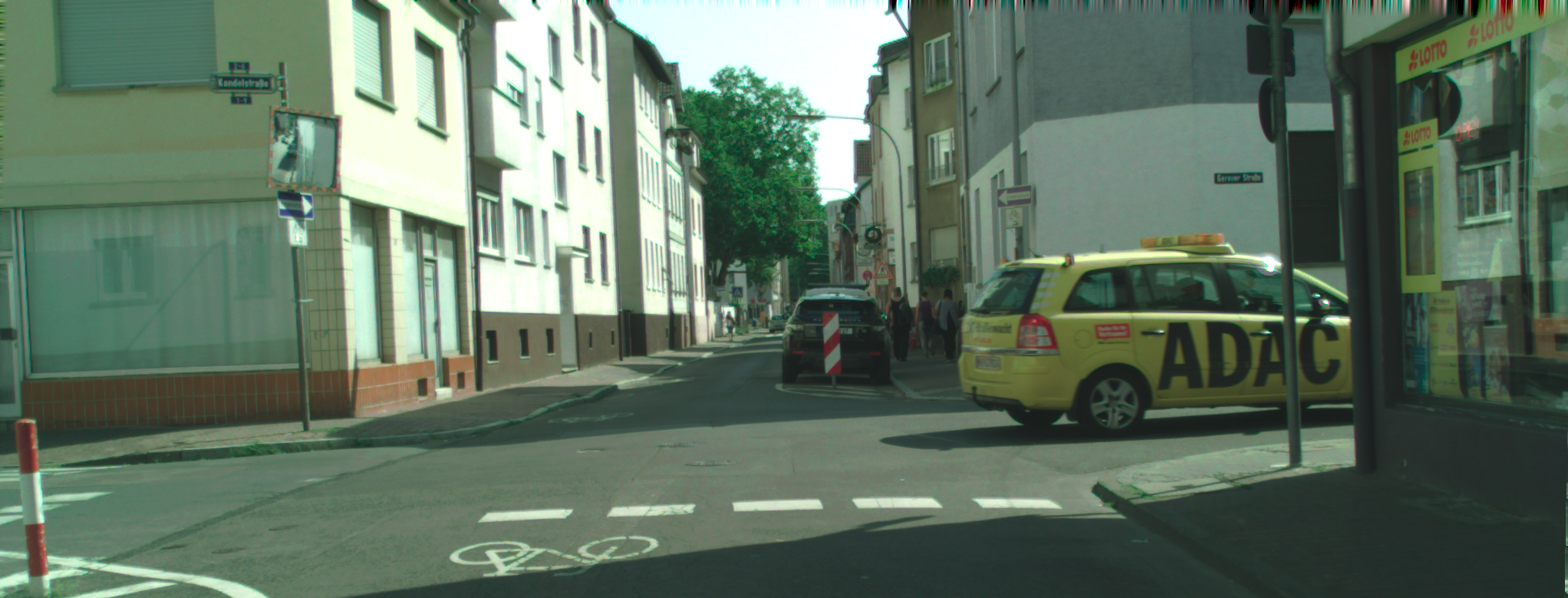}
        \caption{Cityscapes (original) image}
        \label{fig:subfig_a}
    \end{subfigure}
    \begin{subfigure}[t]{0.33\textwidth}
        \centering
        \includegraphics[width=\linewidth]{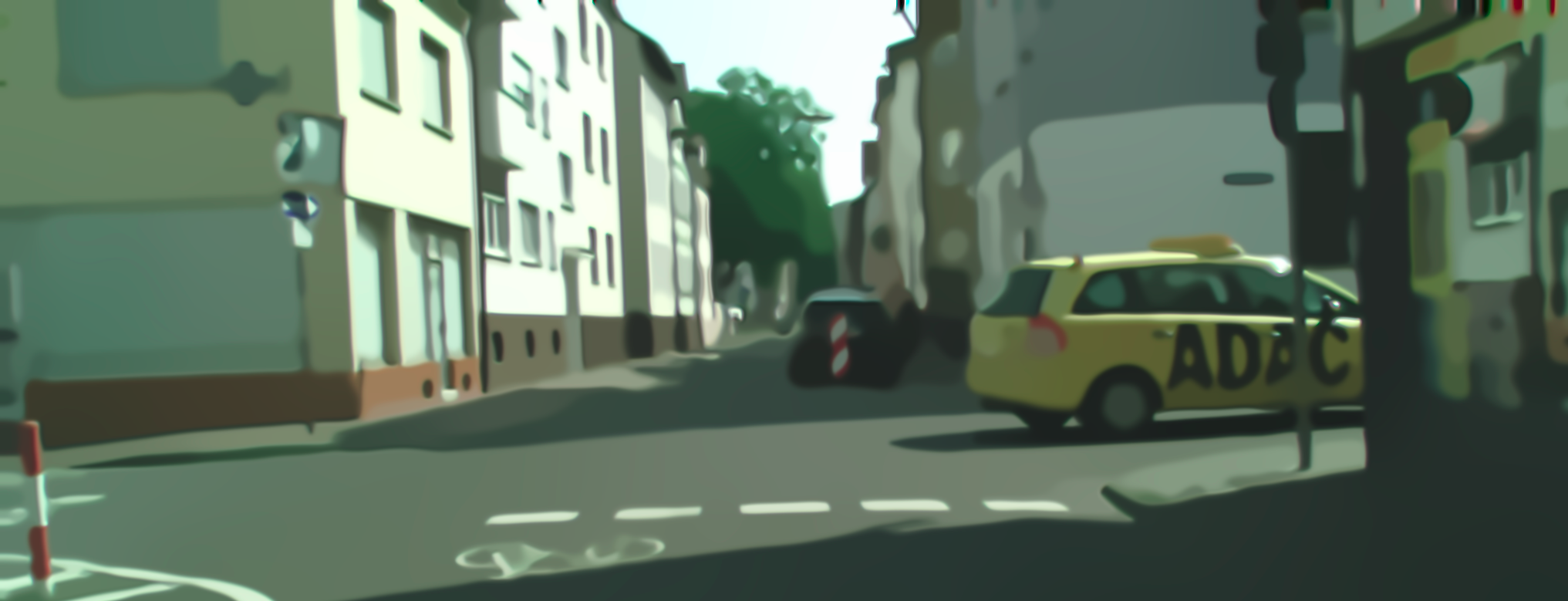}
        \caption{$\mathrm{EED}(\mathrm{City},P_{\mathit{strong}}, 5792)$ image}
        \label{fig:subfig_b}
    \end{subfigure}
    \begin{subfigure}[t]{0.33\textwidth}
        \centering
        \includegraphics[width=\linewidth]{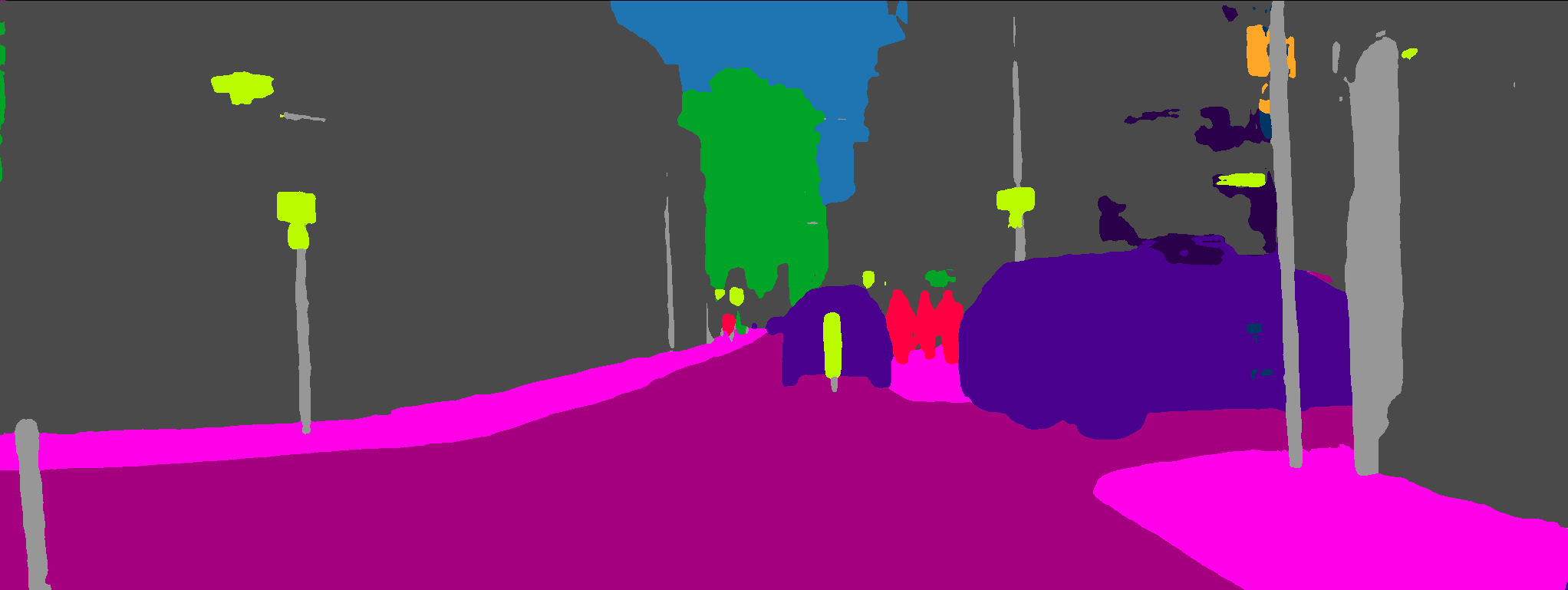}
        \caption{Trained and inferred on Cityscapes }
        \label{fig:subfig_c}
    \end{subfigure}
    \begin{subfigure}[t]{0.33\textwidth}
        \centering
        \includegraphics[width=\linewidth]{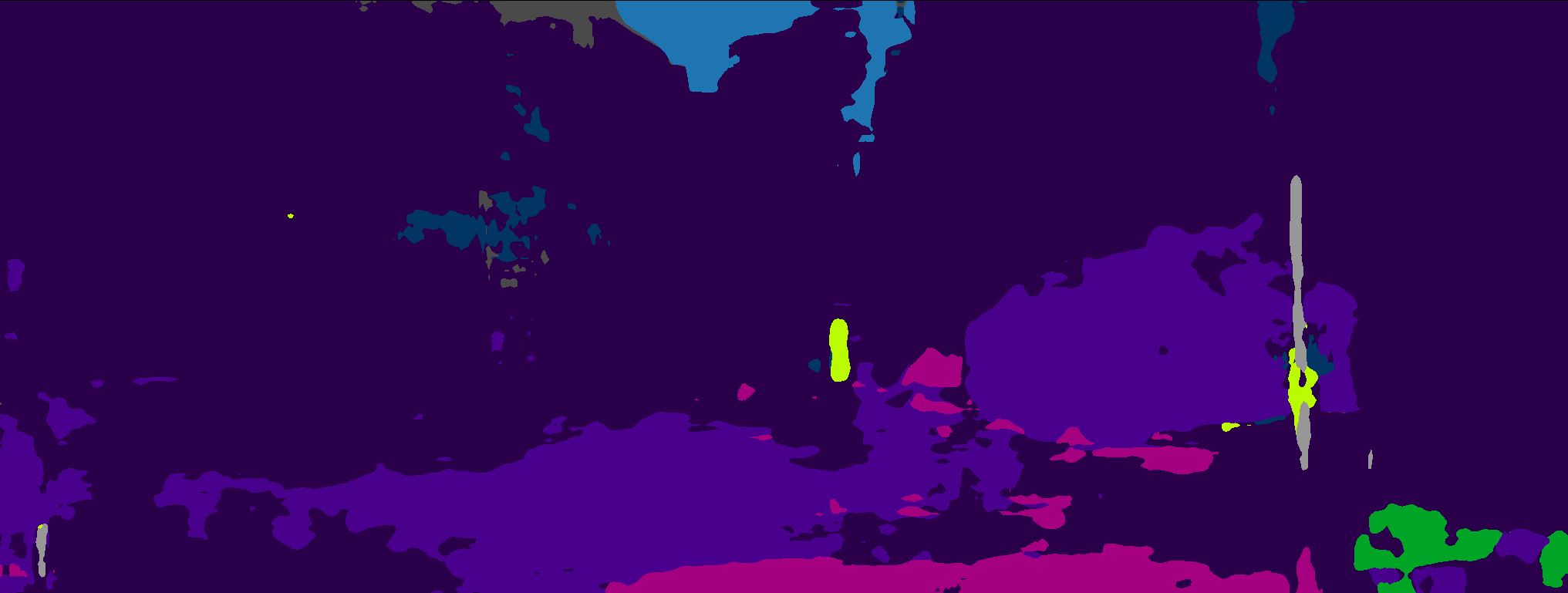}
        \caption{Trained on Cityscapes, inferred on $\mathrm{EED}(\mathrm{City},P_{\mathit{strong}}, 5792)$}
        \label{fig:subfig_d}
    \end{subfigure}

    \medskip

    \begin{subfigure}[t]{0.33\textwidth}
        \centering
        \includegraphics[width=\linewidth]{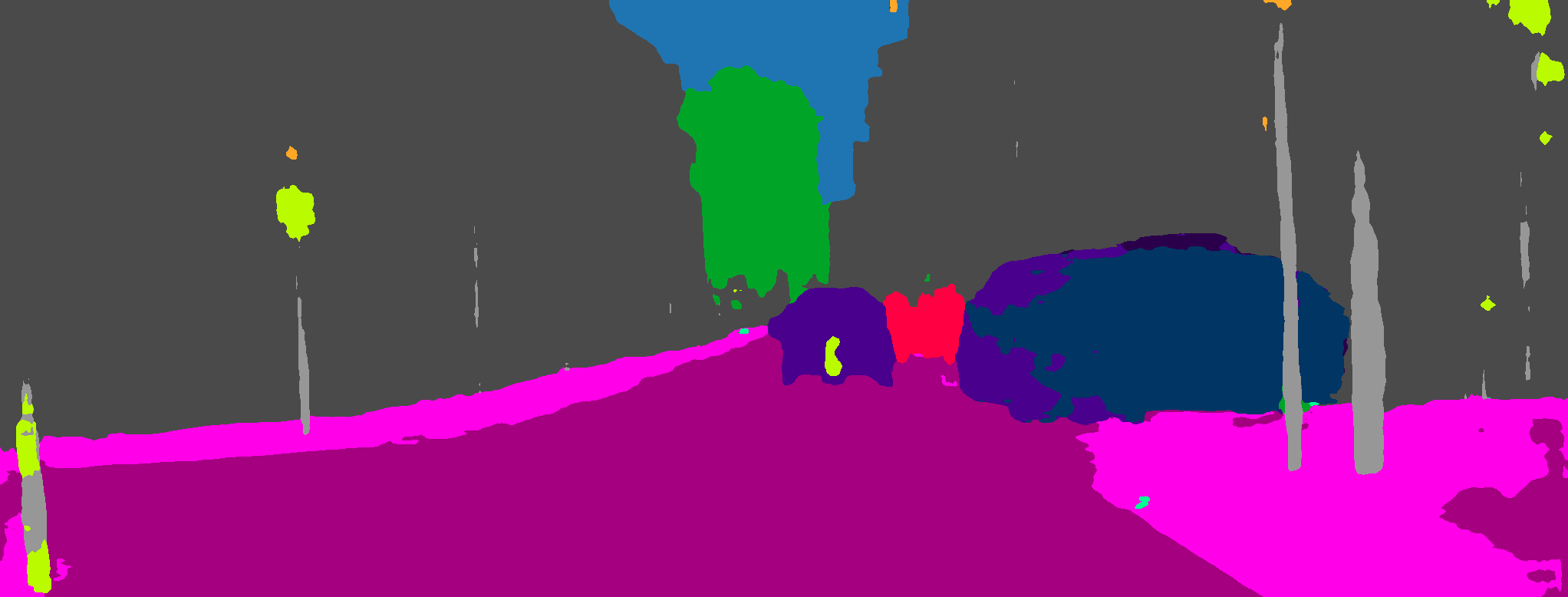}
        \caption{Trained on $\mathrm{EED}(\mathrm{City},P_{\mathit{strong}}, 5792)$, inferred on Cityscapes}
        \label{fig:subfig_e}
    \end{subfigure}
    \begin{subfigure}[t]{0.33\textwidth}
        \centering
        \includegraphics[width=\linewidth]{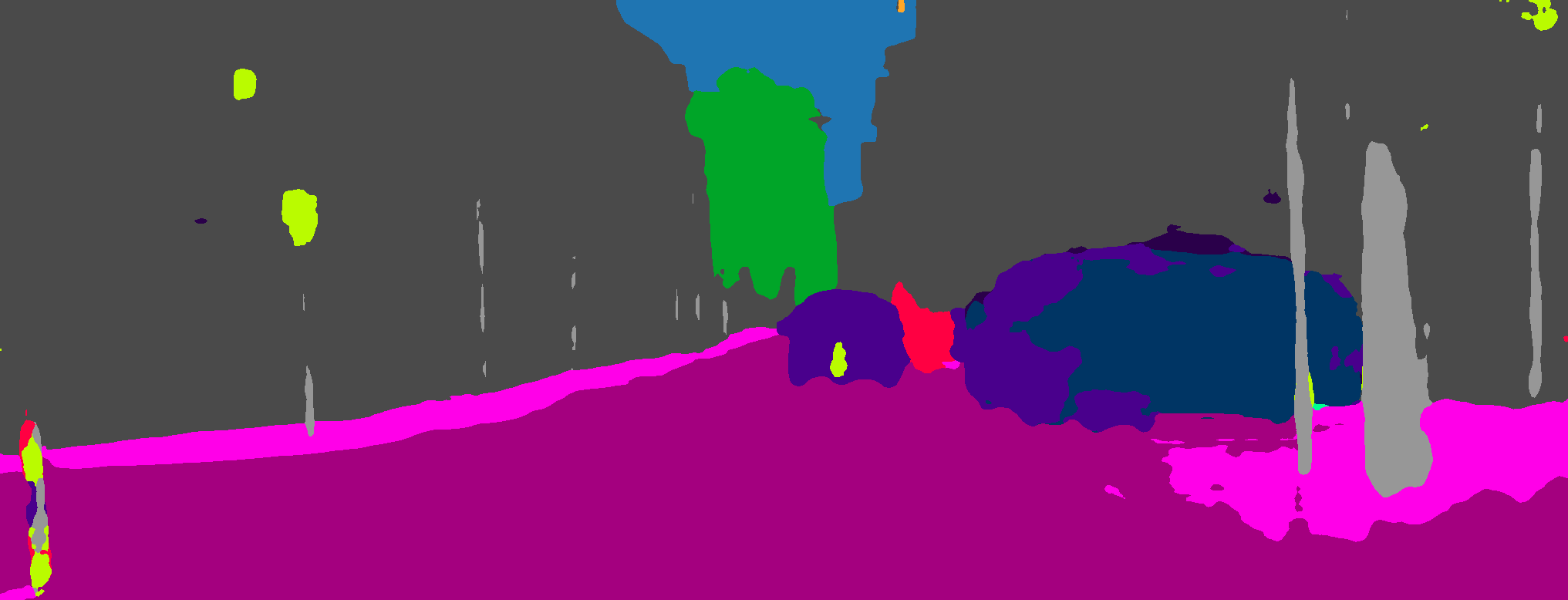}
        \caption{Trained and inferred on $\mathrm{EED}(\mathrm{City},P_{\mathit{strong}}, 5792)$}
        \label{fig:subfig_f}
    \end{subfigure}

    \medskip

    \begin{subfigure}[t]{0.33\textwidth}
        \centering
        \includegraphics[width=\linewidth]{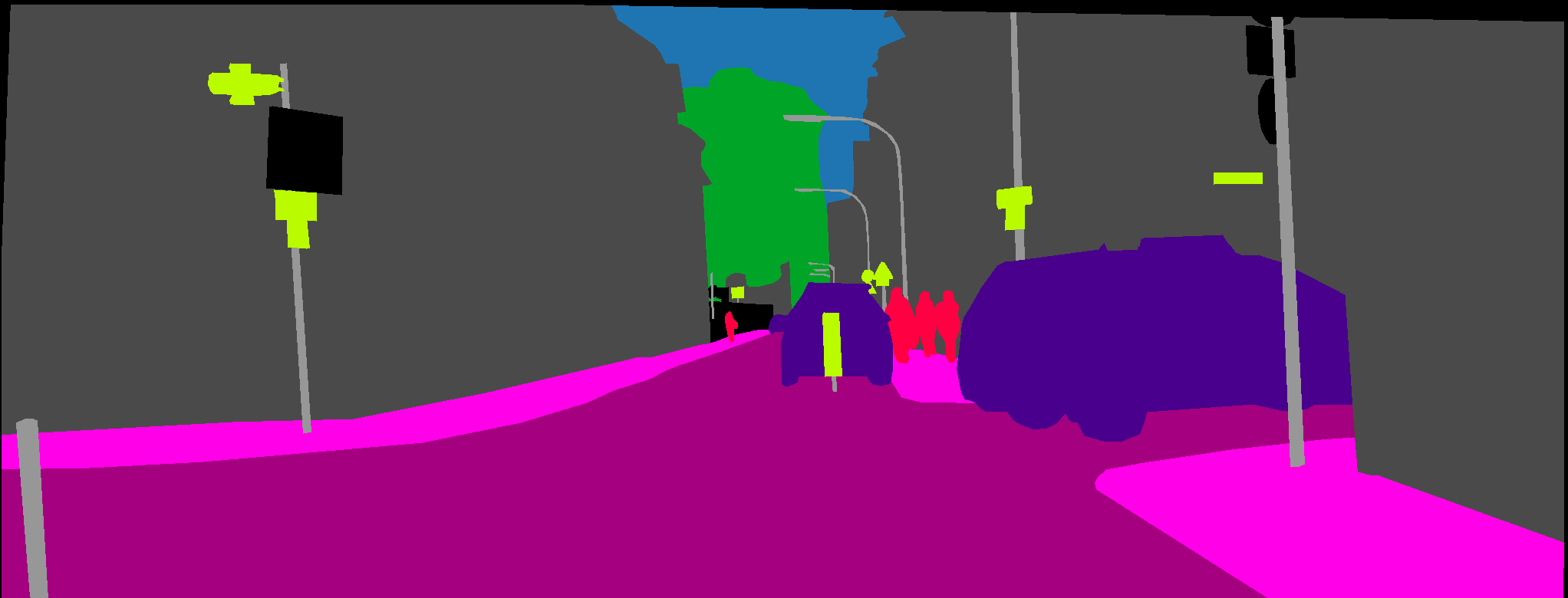}
        \caption{Cityscapes ground truth mask}
        \label{fig:subfig_g}
    \end{subfigure}
    \begin{subfigure}[t]{0.33\textwidth}
        \centering
        \includegraphics[width=\linewidth]{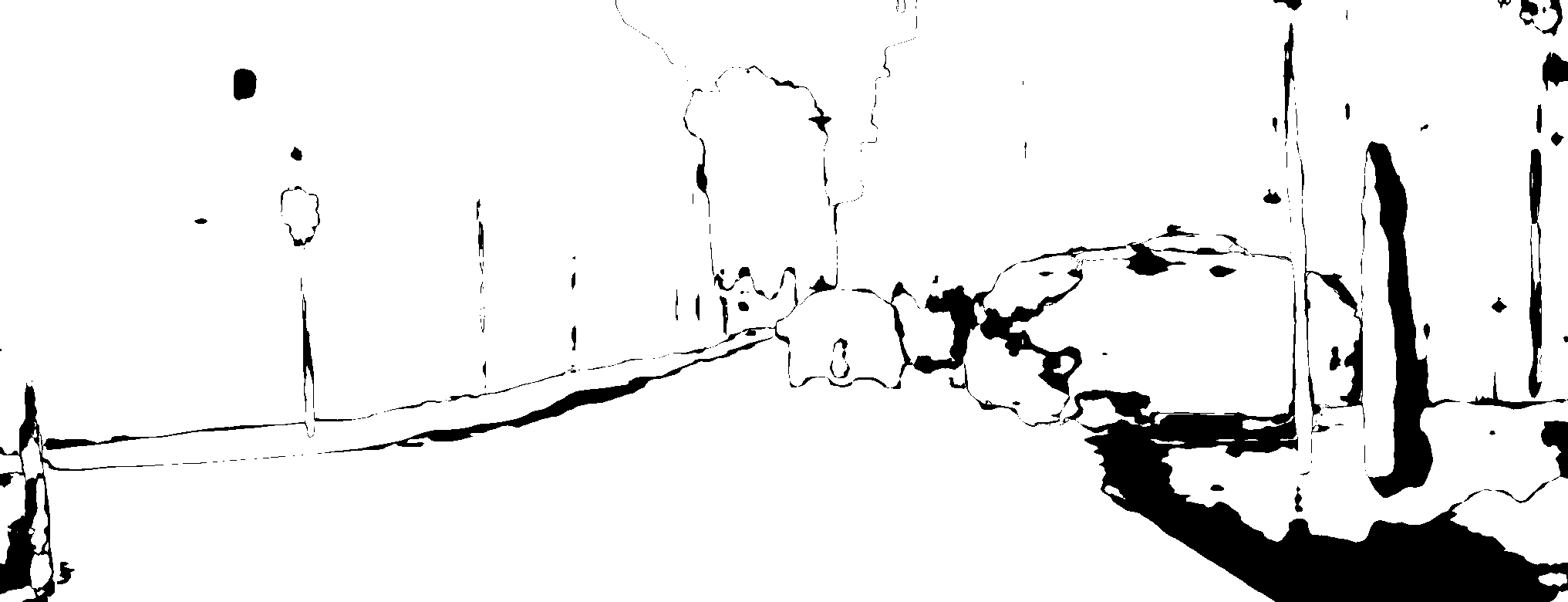}
        \caption{Trained on $P_{\mathit{strong}}, 5792)$, difference of inferences on Cityscapes and $P_{\mathit{strong}}, 5792)$; black pixel means different prediction, white pixel means same prediction}
        \label{fig:subfig_h}
    \end{subfigure}

    \caption{Visual comparison of DeepLabv3+ predictions for different combinations of training data and inferred data.}
    \label{fig:prediction_comparison}
\end{figure*}

\textbf{Semantic segmentation data.} In addition to the original images, we use EED to generate texture-reduced duplicates of the two datasets. We fix the EED parameters based on extensive numerical experiments where we optimize the visual effect of EED w.r.t.\ removing texture while maintaining shape. We chose the contrast parameter $\kappa=1/10$, Gaussian kernel size $9$ and standard deviation $\sigma=3$ to create EED-processed counterparts of both Cityscapes and CARLA. From now on we denote this parameter set by $P_{\mathit{strong}}$. In a later attempt to minimize the changes to high level shape features, i.e., aiming at aligning object contours and label segment borders, we additionally generate counterparts of Cityscapes using the contrast parameter $\kappa=1/15$, Gaussian kernel size $5$ and standard deviation $\sqrt{5}$. We denote this parameter set by $P_{\mathit{mild}}$. The EED-processed datasets are from now on referred to as $\mathrm{EED}(\mathrm{dataset},P, t)$ with $t$ being the number of time steps and $P \in \{P_{\mathit{mild}},P_{\mathit{strong}\}}$. Exemplary images are given in \cref{fig:city_diff} and \cref{fig:prediction_comparison}.

\textbf{Image classification data.} For our proof of concept study in image classification, we also derive classification datasets from Cityscapes and $\mathrm{EED}(\mathrm{City},P_{\mathit{mild}}, 5792)$ by cropping out bounding boxes around segments and using the eleven classes that are suitable for image classification (bicycle, bus, fence, traffic light, truck, wall, building, car, person, traffic sign, vegetation).

\subsection{Image Classification Experiments} \label{sec:imgclassexp}
In our image classification experiments, we train ResNet34 CNNs \citep{he2016deep,khosla2020supervised} on the classification datasets derived from Cityscapes as well as a diffused counterpart $\mathrm{EED}(\mathrm{City},P_{\mathit{mild}}, 5792)$. In \cref{tab:classification}, we perform a comparison by training CNNs on Cityscapes as well as on $(\mathrm{City},P_{\mathit{mild}}, 5792)$ and evaluating the performance of each of the CNNs on both datasets. The classification performance of the CNNs is reported in terms of classification accuracy and balanced accuracy. On the diagonal of the table, we report the test performance of each CNN when trained with a given dataset (Cityscapes and $(\mathrm{City},P_{\mathit{mild}}, 5792)$) and evaluated on a corresponding test set of the same kind. Each number is a mean result over three networks trained with differently initialized weights. A moderate performance decrease can be observed when comparing the top left with the bottom right entry of \cref{tab:classification}, i.e., when comparing CNNs trained and evaluated on Cityscapes with CNNs trained and evaluated on $(\mathrm{City},P_{\mathit{mild}}, 5792)$. This result confirms the texture-dependence of CNNs as reported in the literature, but also shows that quite decent performance can be achieved when training on texture reduced data. The table's off-diagonal presents results on the texture-dependence of CNNs. Clearly, networks trained on original Cityscapes have learned to focus heavily on the texture and use it extensively for decision-making. This is shown by the pronounced drop of around 35 percent points (pp.) when evaluating the network trained with Cityscapes on texture-reduced data from $(\mathrm{City},P_{\mathit{mild}}, 5792)$. On the contrary, CNNs trained on $(\mathrm{City},P_{\mathit{mild}}, 5792)$ have learned to mostly ignore texture. The re-introduction of texture does not lead to confusion of the CNNs, but even to an increase of performance by roughly 2 pp.

\begin{table}[h]
\caption{A study of texture-dependence, based on original and diffused image classification datasets, presented in terms of classification accuracy.}
\scalebox{0.95}{
\begin{tabular}{ |l|c|c| }
\hline
\multicolumn{1}{|c|}{\diagbox{Training}{Evaluation}} & \multicolumn{2}{c|}{Accuracy (Balanced Accuracy)} \\
\cline{2-3}
 & Cityscapes & $\mathrm{EED}(\mathrm{City},P_{\mathit{mild}}, 5792)$ \\
\hline
Cityscapes & \textbf{91.25} (84.96) & 56.52 (50.95) \\
$\mathrm{EED}(\mathrm{City},P_{\mathit{mild}}, 5792)$ & 83.50 (71.62) & \textbf{81.60} (71.38) \\
\hline
\end{tabular}}
\label{tab:classification}
\end{table}

\FloatBarrier
\subsection{Semantic Segmentation Experiments}
The focus of our numerical experiments is on semantic segmentation. For the remainder of this section, we conduct an in-depth study on texture-dependence and texture robustness of ordinarily trained DNNs as well as DNNs trained with EED-processed data. This evaluation contains results for Cityscapes and CARLA as well as for CNNs and Transformers. For our semantic segmentation experiments, we use the MMSegmentation \cite{mmseg2020} framework to train Deeplabv3+ \cite{b20} CNNs from scratch on Cityscapes, $\mathrm{EED}(\mathrm{City},P_{\mathit{strong}}, 5792)$, $\mathrm{EED}(\mathrm{City},P_{\mathit{mild}}, 5792)$, CARLA and $\mathrm{EED}(\mathrm{CARLA},P_{\mathit{mild}}, 5792)$. For each experiment we train three CNNs and report the average result. We refrain from reporting standard errors as they were negligibly small.

\begin{table*}
\centering
\caption{Performance in mIoU of Deeplabv3+ networks trained and evaluated on original and EED-processed Cityscapes and CARLA datasets. For Cityscapes + RandomEED we randomly present the network images from original and EED data during training, where each image is chosen with an 80\% chance from Cityscapes and 5\% from each, $\mathrm{EED}(\mathrm{City},A,B)$ with $(A,B) \in \{ (P_{\mathit{mild}},2896),(P_{\mathit{mild}},8192),(P_{\mathit{strong}},1024),(P_{\mathit{strong}},5792) \}$. }
\scalebox{0.93}{
\begin{tabular}{|l|l l l|l l|}
\hline
\diagbox{trained on}{evaluated on} & Cityscapes & $\mathrm{EED}(\mathrm{City},P_{\mathit{mild}}, 5792)$ & $\mathrm{EED}(\mathrm{City},P_{\mathit{strong}}, 5792)$ & CARLA & $\mathrm{EED}(\mathrm{CARLA},P_{\mathit{strong}}, 5792)$  \\ \hline
Cityscapes & \textbf{72.59} & 19.54 & 9.55 & \textcolor{gray}{48.89} & \textcolor{gray}{11.08} \\
$\mathrm{EED}(\mathrm{City},P_{\mathit{mild}}, 5792)$ & 57.92 & \textbf{56.59} & 50.98 & \textcolor{gray}{32.53} & \textcolor{gray}{25.58} \\
$\mathrm{EED}(\mathrm{City},P_{\mathit{strong}}, 5792)$ & 51.55 & 50.48 & \textbf{47.56} & \textcolor{gray}{24.49} & \textcolor{gray}{19.68} \\ \hline
CARLA & \textcolor{gray}{19.45} & \textcolor{gray}{10.56} & \textcolor{gray}{7.17} & \textbf{78.01} & 31.58 \\
$\mathrm{EED}(\mathrm{CARLA},P_{\mathit{strong}}, 5792)$ & \textcolor{gray}{10.61} & \textcolor{gray}{9.75} & \textcolor{gray}{9.19} & 70.10 & \textbf{70.84} \\ \hline
Cityscapes + RandomEED & 70.15 & 54.45 & 47.22 & \textcolor{gray}{44.45} & \textcolor{gray}{29.66} \\ \hline
\end{tabular}}
\label{tab:cross_evaluation}
\vspace{-10pt}
\end{table*}

\textbf{Comparison of texture-dependence of CNNs.}
We present results of experiments analogous to \cref{sec:imgclassexp}. However, this time we do not report accuracy but rather mIoU. In addition, we present results for Cityscapes and CARLA as well as two different EED configurations, $\mathrm{EED}(\mathrm{City},P_{\mathit{strong}}, 5792)$ and $\mathrm{EED}(\mathrm{City},P_{\mathit{mild}}, 5792)$, for Cityscapes.

The results are summarized in \cref{tab:cross_evaluation}. Qualitatively, we observe similar results compared to the classification experiments, but the effects are much more pronounced. CNNs that have been trained on original data achieve the strongest results in their domain (Cityscapes/CARLA). However, when evaluating those CNNs on EED data, the performance drop is extreme. E.g.\ the CNNs trained on Cityscapes obtain 74.03\% mIoU and this number drops to 8.50\% when evaluating those CNNs on $\mathrm{EED}(\mathrm{City},P_{\mathit{strong}}, 5792)$. On the other hand, CNNs trained on the EED data do not exhibit any performance drop when evaluated on original data. These CNNs successfully learned to ignore texture. A visual study is provided in \cref{fig:prediction_comparison}.

It might be an obvious question whether networks trained on EED data also show stronger domain generalization. The conjecture might be that texture in Cityscapes and CARLA is quite different and might mislead CNNs. However it turns out that it is the other way round. For a given CNN, texture information from Cityscapes seems to be valuable in order to function on CARLA and vice versa. Taking a closer look at data from CARLA, it stands out that the object shapes in CARLA are quite angular compared to Cityscapes. Hence, there is not only a domain shift in texture but also a domain shift in shape, which hinders domain generalization by texture ignorance.

The comparison of the two different EED configurations on Cityscapes in \cref{tab:cross_evaluation} reveals that the ``milder'' EED configuration $\mathrm{EED}(\mathrm{City},P_{\mathit{mild}}, 5792)$ yields better results on Cityscapes and achieves texture ignorance just like the stronger version. The configuration $\mathrm{EED}(\mathrm{City},P_{\mathit{mild}}, 5792)$ yields similar diffusion compared to the stronger configuration within objects, but it shows stronger shape preservation. This improved performance carries over to Cityscapes which indicates that stronger shape preservation is helpful for the learning process as the segmentation masks fit the object shapes. We inspect the importance of the shape fit and visibility of object borders more closely in the upcoming segment-wise analysis.

For Cityscapes + RandomEED we randomly present the network images from original and EED data during training, where each image is chosen with an 80\% chance from Cityscapes and 5\% from a variety of EED configurations, i.e., $\mathrm{EED}(\mathrm{City},(A,B))$ with $(A,B) \in \{ (P_{\mathit{mild}},2896),(P_{\mathit{mild}},8192),(P_{\mathit{strong}},1024),(P_{\mathit{strong}},5792) \}$. This choice of parameters yields an mIoU value on Cityscapes close to the CNN trained on Cityscapes data while also achieving mIoU values on the EED datasets close to the mIoU of the CNNs trained on them, thus being able to operate also when texture is missing.

\begin{figure}
    \includegraphics[width=0.5\textwidth]{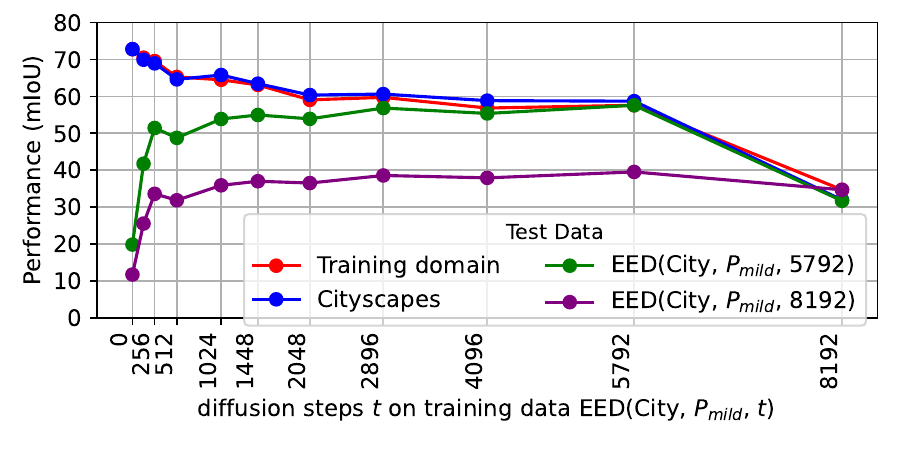}
    \caption{An Ablation study on the effect of the diffusion strength of the training data on the performance on differently diffused test sets. For each value $t$ on the x-axis, a CNN is trained with configuration $\mathrm{EED}(\mathrm{City},P_{\mathit{mild}}, t)$. Each CNN is evaluated on the four different datasets.}
    \label{fig:ablation_study}
\end{figure}

\FloatBarrier
\textbf{Ablation of diffusion steps.} In \cref{fig:ablation_study} we provide an ablation study on the effect of diffusion strength of the training data on the CNN performance on differently diffused test sets. We vary the diffusion strength by considering CNNs trained on $\mathrm{EED}(\mathrm{City},P_{\mathit{mild}}, t)$ for different values of $t$. Each of these CNNs is evaluated on four different datasets, i.e., on a test set from the original Cityscapes as well as from $\mathrm{EED}(\mathrm{City},P_{\mathit{mild}}, k)$ for $k=5792$, $k=8192$ and $k=t$ (equal test diffusion $k$ and training diffusion $t$). The ablation study reveals that when training on $\mathrm{EED}(\mathrm{City},P_{\mathit{mild}}, t)$, texture ignorance is achieved for data $\mathrm{EED}(\mathrm{City},P_{\mathit{mild}}, k)$ with $k \leq t$. For stronger diffusion $k > t$, the performance still degrades. The latter effect becomes stronger as the difference between $k$ and $t$ increases. The CNNs achieve abstraction w.r.t.\ richer but not poorer texture. Notably, e.g., a CNN trained on $\mathrm{EED}(\mathrm{City},P_{\mathit{mild}}, 1024)$ is quite robust w.r.t.\ stronger diffusion from $\mathrm{EED}(\mathrm{City},P_{\mathit{mild}}, 5792)$ while still maintaining most of the mIoU performance on original Cityscapes data.

\begin{table*}
\centering
\caption{A study of texture-dependence for Segformer b1 \cite{xie2021segformer} transformer architecture, 3 runs each, on $\mathrm{EED}(\mathrm{City},P_{\mathit{mild}}, 5792)$ and $\mathrm{EED}(\mathrm{City},P_{\mathit{strong}}, 5792)$.}
\scalebox{0.95}{
\begin{tabular}{|l|c|c|c|}
\hline
\diagbox{Trained on}{Evaluated on} & Cityscapes & $\mathrm{EED}(\mathrm{City},P_{\mathit{mild}}, 5792)$ & $\mathrm{EED}(\mathrm{City},P_{\mathit{strong}}, 5792)$ \\
\hline
Cityscapes & \textbf{63.52} & 39.27 & 26.51 \\
$\mathrm{EED}(\mathrm{City},P_{\mathit{mild}}, 5792)$ & 52.76 & \textbf{52.90} & 50.19 \\
$\mathrm{EED}(\mathrm{City},P_{\mathit{strong}}, 5792)$ & 47.81 & 48.96 & \textbf{47.36} \\
\hline
\end{tabular}}
\label{tab:transformer}
\vspace{-2pt}
\end{table*}

\textbf{Comparison of texture-dependence for transformers.} We now repeat parts of the CNN experiments in \cref{tab:transformer} for the Segformer b1 transformer architecture \cite{xie2021segformer}, training three networks from scratch on Cityscapes, $\mathrm{EED}(\mathrm{City},P_{\mathit{mild}}, 5792)$ and $\mathrm{EED}(\mathrm{City},P_{\mathit{strong}}, 5792)$, respectively. Although there is a noticeable decrease in overall performance--which we attribute to the limited dataset size of 2975 images being insufficient for transformers to achieve their full potential--we observe very similar effects as in our experiments with the DeepLabV3+ architecture. However, in comparison to the CNN case, the transformer model trained on Cityscapes exhibits a far smaller performance loss when evaluated on EED data. Note that this is another confirmation of the texture robustness of transformers that is reported in the literature \cite{Zhang2021Delving}. These results also suggest that one can use EED for broader texture bias evaluation.

\begin{figure*}
    \vspace{-10pt}
    \centering
    \begin{minipage}{0.47\textwidth}
        \includegraphics[trim=0 0 0 1cm, clip, width=\linewidth]{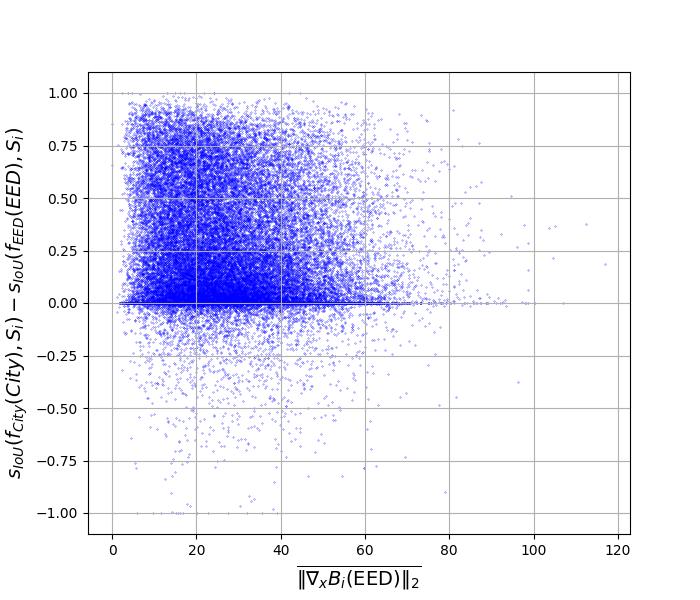}
        \caption{The segmentation performance of $f_{\mathrm{EED}}(\mathrm{EED})$ in comparison to $f_{\mathrm{City}}(\mathrm{City})$ as a function of the visibility of segment boundaries. The smaller $\overline{ \| \nabla_x B_i(\mathrm{EED}) \|_2 } $, the less visible the segment boundary.}
        \label{fig:first_image}
    \end{minipage}\hfill
    \begin{minipage}{0.47\textwidth}
        \includegraphics[trim=0 0 0 1cm, clip, width=\linewidth]{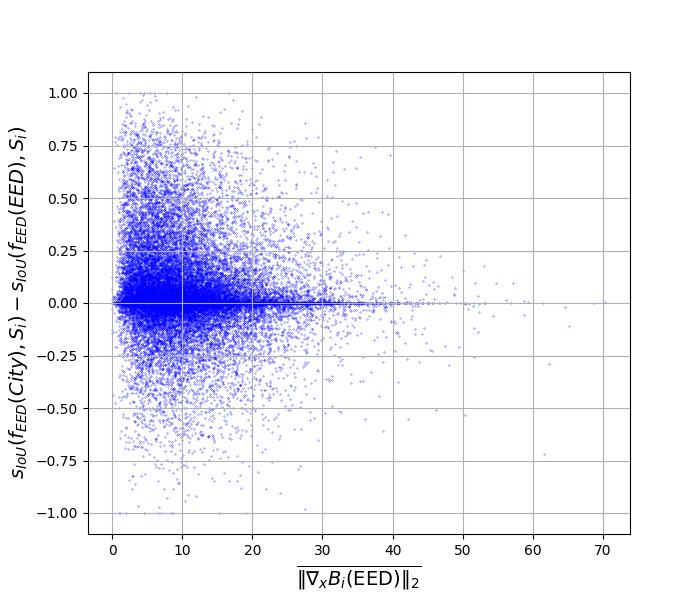}
        \caption{The segmentation performance of $f_{\mathrm{EED}}(EED)$ in comparison to $f_{\mathrm{EED}}(\mathrm{City})$ as a function of the visibility of segment boundaries. The smaller $\overline{ \| \nabla_x B_i(\mathrm{EED}) \|_2 } $, the less visible the segment boundary.}
        \label{fig:second_image}
    \end{minipage}
\end{figure*}

\textbf{Segment-wise analysis.} In this analysis and the subsequent one, we study two effects in more detail: 
\begin{itemize}
    \item[E1)] The performance loss when training and evaluating on Cityscapes vs.\ when training and evaluation on EED data.
    \item[E2)] The almost equal performance when training on EED data and evaluating on both Cityscapes and EED data.
\end{itemize}
Utilizing the Metaseg analysis tool \cite{rottmann2020prediction}, we conduct an analysis looking at the prediction accuracy on each segment and thereafter we consider class-wise (global) IoU values over the whole dataset (of which the class average gives the regular mIoU). In the segment-wise and the class-wise analysis we use the shorthand $\mathrm{EED}$ for $\mathrm{EED}(\mathrm{City},P_{\mathit{strong}}, 5792)$.

Every image contains a number of ground truth segments $S_i$, $i=1,\ldots,m$. Let $B_i$ denote the boundary pixels of the $i$th segment. One could image that part of the performance loss is due to vanishing object boundaries caused by over-diffusion. To study this effect, we compute gradients in boundary pixels of $B_i$ w.r.t.\ the image domain $x$, and average their euclidean norms. Let $B_i(\mathrm{EED})$ denote a ground truth segment's boundary pixels of an EED-processed image and $B_i(\mathrm{City})$ has the obvious analogous meaning w.r.t.\ Cityscapes. Then the desired quantity is given by $ \overline{ \| \nabla_x B_i(\mathrm{EED}) \|_2 } $. To measure the segment-wise performance of a given CNN $f_{\mathrm{EED}}$ trained on EED-processed data (analogously we can consider $f_{\mathrm{City}}$), we use a segment-wise IoU, termed $s_{\mathrm{IoU}}$, that computes the segment-wise IoU of a ground truth segment $S_i$ and a network's prediction $f_A$, $A \in \{ \mathrm{City}, \mathrm{EED} \}$. The performance loss/gain of $f_{\mathrm{EED}}$ compared to $f_{\mathrm{City}}$ is then quantified by 
$s_{\mathrm{IoU}}( f_{\mathrm{City}}(\mathrm{City}), S_i ) - s_{\mathrm{IoU}}( f_{\mathrm{EED}}(\mathrm{EED}), S_i )$, where $f_{A}(B)$ denotes $f$ trained on $A$ and inferred on $B$, with $A,B \in \{ \mathrm{City}, \mathrm{EED} \}$. To further understand effect E1, we study the connection of the latter quantity to $ \overline{ \| \nabla_x B_i(\mathrm{EED}) \|_2 } $, i.e., how much does the loss of segment boundary image information correlate with change in performance, in \cref{fig:first_image}. Visually, the accuracy in terms of $s_{\mathrm{IoU}}$ of $f_{\mathrm{EED}}(\mathrm{EED})$ and $f_{\mathrm{City}}(\mathrm{City})$ is more similar when the boundary $B_i$ is more pronounced, i.e., when $ \overline{ \| \nabla_x B_i(\mathrm{EED}) \|_2 } $ is higher.

Analogously, we can study part of the effect E2 by considering the correlation of $s_{\mathrm{IoU}}( f_{\mathrm{EED}}(\mathrm{City}), S_i ) - s_{\mathrm{IoU}}( f_{\mathrm{EED}}(\mathrm{EED}), S_i )$ with  $ \overline{ \| \nabla_x B_i(\mathrm{EED}) \|_2 } $ to see how similar the predictions of $f_{\mathrm{EED}}$ on both data sources $\mathrm{City}$ and $\mathrm{EED}$ are, cf.\ \cref{fig:second_image}. Here, the rather centered fluctuations around zero on the y-axis are in agreement with the similar performance of $f_{\mathrm{EED}}(\mathrm{City})$ and $f_{\mathrm{EED}}(\mathrm{EED})$. However, it can be seen that a lack of boundary visibility, i.e., low values of $ \overline{ \| \nabla_x B_i(\mathrm{EED}) \|_2 } $ result in higher variability of the predictions of $f_{\mathrm{EED}}$.

\begin{table*}
\centering
\caption{Class-wise comparison of IoU performances of different combinations of $f_A(B)$, $A,B \in \{ \mathrm{City}, \mathrm{EED} \} $.}
\scalebox{0.95}{
\begin{tabular}{|l|rrrrr|}
\hline
\diagbox{IoU (class)}{Evaluation}    & $f_{\mathrm{City}}(\mathrm{City})$ & $f_{\mathrm{EED}}(\mathrm{EED})$ & $f_{\mathrm{EED}}(\mathrm{City})$ & \makecell{$f_{\mathrm{City}}(\mathrm{City})$ 
 \\ $- f_{\mathrm{EED}}(\mathrm{EED})$} & \makecell{$f_{\mathrm{EED}}(\mathrm{City})$ \\ $ - f_{\mathrm{EED}}(\mathrm{EED})$} \\
\hline
\rowcolor{green!20}
road             & 97.42      & 90.10      & 92.72      & 7.32                    & 2.62                    \\
sidewalk         & 81.88      & 50.32      & 60.65      & 31.56                   & 10.33                   \\
building         & 90.72      & 76.20      & 80.38      & 14.52                   & 4.18                    \\
wall             & 51.54      & 8.85       & 10.74      & 42.69                   & 1.89                    \\
pole             & 62.83      & 34.53      & 40.39      & 28.30                   & 5.86                    \\
\rowcolor{red!15}
traffic light    & 62.61      & 23.95      & 22.98      & 38.66                   & -0.97                   \\
\arrayrulecolor{red!15}\hline \arrayrulecolor{black}
\rowcolor{red!15}
traffic sign     & 74.25      & 38.31      & 41.96      & 35.94                   & 3.65                    \\
vegetation       & 91.46      & 80.70      & 81.67      & 10.76                   & 0.97                    \\
terrain          & 50.25      & 31.30      & 26.97      & 18.95                   & -4.33                   \\
\rowcolor{green!20}
sky              & 94.06      & 89.18      & 87.93      & 4.88                    & -1.25                   \\
person           & 82.16      & 55.07      & 56.15      & 27.09                   & 1.08                    \\
car              & 93.56      & 78.75      & 81.79      & 14.81                   & 3.04                    \\
truck            & 38.78      & 8.16       & 11.34      & 30.62                   & 3.18                    \\
bus              & 64.91      & 38.27      & 40.26      & 26.64                   & 1.99                    \\
\hline
\textbf{Mean} (mIoU)   & 74.03      & 50.26      & 52.57      & 23.77                   & 2.30                    \\
\hline
\end{tabular}}
\label{tab:classwise}
\vspace{-10pt}
\end{table*}

\textbf{Class-wise analysis.} We study the effects E1 and E2 defined in the segment-wise analysis by considering class-wise IoU values to investigate whether certain classes lose performance due to EED-processed training data, cf.\ \cref{tab:classwise}.
Comparing $f_{\mathrm{EED}}(\mathrm{EED})$ and $f_{\mathrm{City}}(\mathrm{City})$, the classes road and sky with large segments exhibit the smallest performance loss. On the other hand, classes with smaller segments like traffic light and traffic sign experience a large performance drop due to EED-processing. Also the IoU on class person reduces stronger than for the class car. Thus, these results are in accordance to our segment-wise analysis. When comparing $f_{\mathrm{EED}}(\mathrm{EED})$ and $f_{\mathrm{EED}}(\mathrm{City})$, one can observe that in particular the classes sidewalk and pole show the strongest difference in performance and can be perceived better when presenting original data to $f_{\mathrm{EED}}$. When looking into diffused images, we indeed observed that poles and sidewalk borders are often difficult to perceive.

We conclude that the clear visibility of object borders is of importance for CNNs, in particular when texture is missing, and that EED, as to be expected, makes smaller objects difficult to perceive for CNNs.

\textbf{Robustness against adversarial attacks.}
An obvious question might be, whether EED-based pre-processing also increases adversarial robustness, i.e., robustness against adversarial attacks. Adversarial attacks add small perturbations to the input image leading the neural network to erroneous predictions during testing \citep{Bar2021,Maag_2024_WACV}. These perturbations are not perceptible to humans, making the robustness against these examples highly relevant. To this end, detection methods have been developed which distinguish between clean and perturbed inputs \citep{Maag2023,Xiao2018} as well as defense methods which increase robustness, making it more difficult to generate adversarial examples \citep{Arnab2018,Xu2021}. 
In the following, we apply adversarial attacks to DNNs trained on original data and to DNNs trained on EED-processed data and compare the robustness of these networks.
The performance of adversarial attackers is evaluated by the attack pixel success rate \cite{Rony2022} which is equivalent to $1-\text{ACC}$ (accuracy). Since the mIoU values obtained for Cityscapes differ for standard and diffusion-based training, we introduce a relative accuracy metric to ensure a fair comparison of the methods. This metric is defined by
\begin{equation}\label{eq:apsr}
    \text{ACC}_{\mathit{rel}} = 1- \frac{\text{ACC}_{\mathit{CS}} - \text{ACC}_{\mathit{AA}}}{\text{ACC}_{\mathit{CS}}}
\end{equation}
where $\text{ACC}_{\mathit{CS}}$ describes the accuracy of the network evaluated on Cityscapes and $\text{ACC}_{\mathit{AA}}$ after performing the adversarial attack. 
As attack we consider the often used single-step \emph{fast gradient sign method} (FGSM, \cite{Goodfellow2015}) in the untargeted as well as target version where the least likely class predicted by the model is chosen as target class following the convention. We denote the attack by FGSM$_{\varepsilon}^{\#}$ where the magnitude of perturbation is given by $\varepsilon = \{ 2,4,8,16 \} $ and the superscript ($\# \in \{ \_,ll \}$) discriminates between untargeted and targeted ($ll$ refers to "least likely").

\begin{figure}
    \centering
    \includegraphics[width=0.45\textwidth]{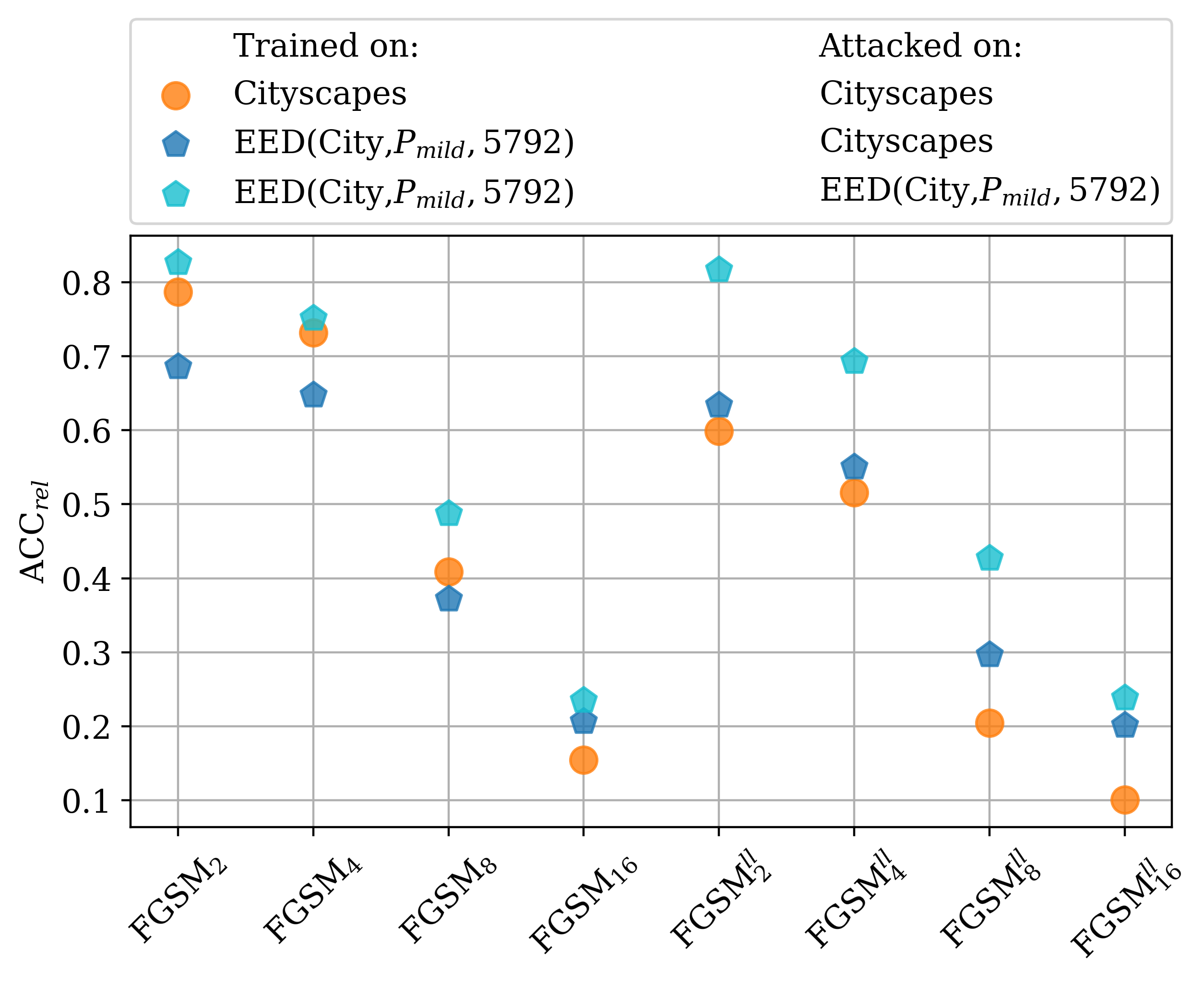}
    \caption{$\text{ACC}_{\mathit{rel}}$ results for the FGSM attacks for CNNs trained and attacked on original data (orange), trained on EED$(\mathrm{City},P_{\mathit{mild}}, 5792)$ and attacked on original data (blue), as well as trained and attacked on EED$(\mathrm{City},P_{\mathit{mild}}, 5792)$ (light blue).}
    \label{fig:attack}
    \vspace{-20pt}
\end{figure}

The numerical results for CNNs with different combinations of Cityscapes and $(\mathrm{City},P_{\mathit{mild}}, 5792)$ datasets for training and attacking, evaluated with respect to the $\text{ACC}_{\mathit{rel}}$ evaluation metric, are given in \cref{fig:attack}. As per usual, the general tendency is that the $\text{ACC}_{\mathit{rel}}$ values are higher for attacks with less magnitude of perturbation. In five cases, i.e., the untargeted case with $\varepsilon=16$ and all targeted cases, the model trained on EED-processed data outperforms the one trained on Cityscapes when providing original Cityscapes images with adversarial attacks to the models. Thus, generally improved adversarial robustness cannot be claimed. However, EED pre-processing can be used to defend against adversarial attacks. It can be expected that EED filters adversarial attacks. We consider the case where an attacker has access to the images after EED-processing, before they go into the CNN. When attacking EED-processed images, we still obtain clearly improved results compared to CNNs trained and attacked w.r.t.\ original data (\cref{fig:attack}, light blue), demonstrating the potential of EED for adversarial defense.

\section{Conclusion}

In this work we utilize EED as a pre-processing to study the texture bias of semantic segmentation models, including CNNs and vision transformers, as well as of classification models. By means of EED-processed duplicates of Cityscapes and a dataset extracted from the CARLA driving simulation, we were able to show the significant texture-dependence of these models. All our models have been trained from scratch. By training DNNs on EED-processed images, we achieve ignorance of those DNNs w.r.t.\ local textural patterns. A detailed analysis on segment level reveals that most of the performance loss on EED-processed images can be attributed to over-diffused cases where shape information is lost. Online EED pre-processing can help to reduce the effect of adversarial attacks. Although one might expect better domain generalization when domains mostly differ w.r.t.\ texture, our analysis on Cityscapes and CARLA data reveals that this setup also contains a significant non-texture-related domain gap. We expect that EED can serve more generally as part of an evaluation protocoll for texture bias reduction methods, where different networks are compared on EED-processed and original datasets to evaluate the degree of texture bias of the networks.

\section*{Acknowledgment}

We would like to acknowledge T.\ Gottwald for conducting image classification experiments. E.\ H.\ \& M.\ R. acknowledge fruitful discussions with H.\ Gottschalk and A.\ M\"utze. E.H.\ and M.R.\ acknowledge support by the German Federal Ministry of Education and Research within the junior research group project “UnrEAL” (grant no.\ 01IS22069).

\bibliographystyle{plain}
\bibliography{literature}

\begin{thebibliography}{10}

\bibitem{Arnab2018}
Anurag Arnab, Ondrej Miksik, and Philip Torr.
\newblock On the robustness of semantic segmentation models to adversarial attacks.
\newblock In {\em IEEE/CVF Conference on Computer Vision and Pattern Recognition (CVPR)}, 2018.

\bibitem{Bar2021}
Andreas Bar, Jonas Lohdefink, Nikhil Kapoor, Serin Varghese, Fabian Huger, Peter Schlicht, and Tim Fingscheidt.
\newblock The vulnerability of semantic segmentation networks to adversarial attacks in autonomous driving: Enhancing extensive environment sensing.
\newblock {\em IEEE Signal Processing Magazine}, 2021.

\bibitem{b10}
David Bau, Bolei Zhou, Aditya Khosla, Aude Oliva, and Antonio Torralba.
\newblock Network dissection: Quantifying interpretability of deep visual representations.
\newblock In {\em Proceedings of the IEEE conference on computer vision and pattern recognition}, pages 6541--6549, 2017.

\bibitem{charbonnier1997deterministic}
Pierre Charbonnier, Laure Blanc-F{\'e}raud, Gilles Aubert, and Michel Barlaud.
\newblock Deterministic edge-preserving regularization in computed imaging.
\newblock {\em IEEE Transactions on image processing}, 6(2):298--311, 1997.

\bibitem{b20}
Liang-Chieh Chen, Yukun Zhu, George Papandreou, Florian Schroff, and Hartwig Adam.
\newblock Encoder-decoder with atrous separable convolution for semantic image segmentation.
\newblock In {\em Proceedings of the European conference on computer vision (ECCV)}, pages 801--818, 2018.

\bibitem{mmseg2020}
MMSegmentation Contributors.
\newblock {MMSegmentation}: Openmmlab semantic segmentation toolbox and benchmark.
\newblock \url{https://github.com/open-mmlab/mmsegmentation}, 2020.

\bibitem{b18}
Marius Cordts, Mohamed Omran, Sebastian Ramos, Timo Scharw{\"a}chter, Markus Enzweiler, Rodrigo Benenson, Uwe Franke, Stefan Roth, and Bernt Schiele.
\newblock The cityscapes dataset.
\newblock In {\em CVPR Workshop on the Future of Datasets in Vision}, volume~2. sn, 2015.

\bibitem{b17}
Dawei Dai, Yutang Li, Yuqi Wang, Huanan Bao, and Guoyin Wang.
\newblock Rethinking the image feature biases exhibited by deep convolutional neural network models in image recognition.
\newblock {\em CAAI Transactions on Intelligence Technology}, 7(4):721--731, 2022.

\bibitem{dosovitskiy2020image}
Alexey Dosovitskiy, Lucas Beyer, Alexander Kolesnikov, Dirk Weissenborn, Xiaohua Zhai, Thomas Unterthiner, Mostafa Dehghani, Matthias Minderer, Georg Heigold, Sylvain Gelly, et~al.
\newblock An image is worth 16x16 words: Transformers for image recognition at scale.
\newblock {\em arXiv preprint arXiv:2010.11929}, 2020.

\bibitem{b19}
Alexey Dosovitskiy, German Ros, Felipe Codevilla, Antonio Lopez, and Vladlen Koltun.
\newblock Carla: An open urban driving simulator.
\newblock In {\em Conference on robot learning}, pages 1--16. PMLR, 2017.

\bibitem{b7}
Irena Gali{\'c}, Joachim Weickert, Martin Welk, Andr{\'e}s Bruhn, Alexander Belyaev, and Hans-Peter Seidel.
\newblock Image compression with anisotropic diffusion.
\newblock {\em Journal of Mathematical Imaging and Vision}, 31:255--269, 2008.

\bibitem{b13}
Songwei Ge, Shlok Mishra, Chun-Liang Li, Haohan Wang, and David Jacobs.
\newblock Robust contrastive learning using negative samples with diminished semantics.
\newblock {\em Advances in Neural Information Processing Systems}, 34:27356--27368, 2021.

\bibitem{b11}
Robert Geirhos, Patricia Rubisch, Claudio Michaelis, Matthias Bethge, Felix~A Wichmann, and Wieland Brendel.
\newblock Imagenet-trained cnns are biased towards texture; increasing shape bias improves accuracy and robustness.
\newblock {\em arXiv preprint arXiv:1811.12231}, 2018.

\bibitem{Goodfellow2015}
Ian~J. Goodfellow, Jonathon Shlens, and Christian Szegedy.
\newblock Explaining and harnessing adversarial examples.
\newblock In Yoshua Bengio and Yann LeCun, editors, {\em International Conference on Learning Representations (ICLR)}, 2015.

\bibitem{he2016deep}
Kaiming He, Xiangyu Zhang, Shaoqing Ren, and Jian Sun.
\newblock Deep residual learning for image recognition.
\newblock In {\em Proceedings of the IEEE conference on computer vision and pattern recognition}, pages 770--778, 2016.

\bibitem{b16}
Xilin He, Qinliang Lin, Cheng Luo, Weicheng Xie, Siyang Song, Feng Liu, and Linlin Shen.
\newblock Shift from texture-bias to shape-bias: Edge deformation-based augmentation for robust object recognition.
\newblock In {\em Proceedings of the IEEE/CVF International Conference on Computer Vision}, pages 1526--1535, 2023.

\bibitem{jacob2021qualitative}
Georgin Jacob, RT~Pramod, Harish Katti, and SP~Arun.
\newblock Qualitative similarities and differences in visual object representations between brains and deep networks.
\newblock {\em Nature communications}, 12(1):1872, 2021.

\bibitem{b12}
Dhruva Kashyap, Sumukh~K Aithal, C~Rakshith, and Natarajan Subramanyam.
\newblock Towards domain adversarial methods to mitigate texture bias.
\newblock In {\em ICML 2022: Workshop on Spurious Correlations, Invariance and Stability}, 2022.

\bibitem{khosla2020supervised}
Prannay Khosla, Piotr Teterwak, Chen Wang, Aaron Sarna, Yonglong Tian, Phillip Isola, Aaron Maschinot, Ce~Liu, and Dilip Krishnan.
\newblock Supervised contrastive learning.
\newblock {\em arXiv preprint arXiv:2004.11362}, 2020.

\bibitem{b14}
Myeongjin Kim and Hyeran Byun.
\newblock Learning texture invariant representation for domain adaptation of semantic segmentation.
\newblock In {\em Proceedings of the IEEE/CVF conference on computer vision and pattern recognition}, pages 12975--12984, 2020.

\bibitem{krizhevsky2012imagenet}
Alex Krizhevsky, Ilya Sutskever, and Geoffrey~E Hinton.
\newblock Imagenet classification with deep convolutional neural networks.
\newblock {\em Advances in neural information processing systems}, 25, 2012.

\bibitem{lecun1998gradient}
Yann LeCun, L{\'e}on Bottou, Yoshua Bengio, and Patrick Haffner.
\newblock Gradient-based learning applied to document recognition.
\newblock {\em Proceedings of the IEEE}, 86(11):2278--2324, 1998.

\bibitem{b15}
Sangjun Lee, Inwoo Hwang, Gi-Cheon Kang, and Byoung-Tak Zhang.
\newblock Improving robustness to texture bias via shape-focused augmentation.
\newblock In {\em Proceedings of the IEEE/CVF Conference on Computer Vision and Pattern Recognition}, pages 4323--4331, 2022.

\bibitem{Maag2023}
Kira Maag and Asja Fischer.
\newblock Uncertainty-based detection of adversarial attacks in semantic segmentation.
\newblock {\em ArXiv}, 2023.

\bibitem{Maag_2024_WACV}
Kira Maag and Asja Fischer.
\newblock Uncertainty-weighted loss functions for improved adversarial attacks on semantic segmentation.
\newblock In {\em Proceedings of the IEEE/CVF Winter Conference on Applications of Computer Vision (WACV)}, pages 3906--3914, January 2024.

\bibitem{b8}
Shlok Mishra, Anshul Shah, Ankan Bansal, Janit Anjaria, Jonghyun Choi, Abhinav Shrivastava, Abhishek Sharma, and David Jacobs.
\newblock Learning visual representations for transfer learning by suppressing texture.
\newblock {\em arXiv preprint arXiv:2011.01901}, 2020.

\bibitem{b6}
Pietro Perona and Jitendra Malik.
\newblock Scale-space and edge detection using anisotropic diffusion.
\newblock {\em IEEE Transactions on pattern analysis and machine intelligence}, 12(7):629--639, 1990.

\bibitem{b1}
Pietro Perona, Takahiro Shiota, and Jitendra Malik.
\newblock Anisotropic diffusion.
\newblock {\em Geometry-driven diffusion in computer vision}, pages 73--92, 1994.

\bibitem{b5}
Ilya Pollak, Alan~S Willsky, and Hamid Krim.
\newblock Image segmentation and edge enhancement with stabilized inverse diffusion equations.
\newblock {\em IEEE transactions on image processing}, 9(2):256--266, 2000.

\bibitem{Rony2022}
Jérôme Rony, Jean-Christophe Pesquet, and Ismail~Ben Ayed.
\newblock Proximal splitting adversarial attacks for semantic segmentation, 2022.

\bibitem{rottmann2020prediction}
Matthias Rottmann, Pascal Colling, Thomas~Paul Hack, Robin Chan, Fabian H{\"u}ger, Peter Schlicht, and Hanno Gottschalk.
\newblock Prediction error meta classification in semantic segmentation: Detection via aggregated dispersion measures of softmax probabilities.
\newblock In {\em 2020 International Joint Conference on Neural Networks (IJCNN)}, pages 1--9. IEEE, 2020.

\bibitem{tuli2021convolutional}
Shikhar Tuli, Ishita Dasgupta, Erin Grant, and Thomas~L Griffiths.
\newblock Are convolutional neural networks or transformers more like human vision?
\newblock {\em arXiv preprint arXiv:2105.07197}, 2021.

\bibitem{vaswani2017attention}
Ashish Vaswani, Noam Shazeer, Niki Parmar, Jakob Uszkoreit, Llion Jones, Aidan~N Gomez, {\L}ukasz Kaiser, and Illia Polosukhin.
\newblock Attention is all you need.
\newblock {\em Advances in neural information processing systems}, 30, 2017.

\bibitem{b3}
Joachim Weickert.
\newblock {\em Theoretical foundations of anisotropic diffusion in image processing}.
\newblock Springer, 1996.

\bibitem{b4}
Joachim Weickert.
\newblock A review of nonlinear diffusion filtering.
\newblock In {\em International conference on scale-space theories in computer vision}, pages 1--28. Springer, 1997.

\bibitem{b2}
Joachim Weickert.
\newblock {\em Anisotropic diffusion in image processing}, volume~1.
\newblock Teubner Stuttgart, 1998.

\bibitem{weickert1999coherence}
Joachim Weickert.
\newblock Coherence-enhancing diffusion filtering.
\newblock {\em International journal of computer vision}, 31:111--127, 1999.

\bibitem{weickert2006tensor}
Joachim Weickert and Martin Welk.
\newblock Tensor field interpolation with pdes.
\newblock In {\em Visualization and processing of tensor fields}, pages 315--325. Springer, 2006.

\bibitem{weickert20132}
Joachim Weickert, Martin Welk, and Marco Wickert.
\newblock L 2-stable nonstandard finite differences for anisotropic diffusion.
\newblock In {\em Scale Space and Variational Methods in Computer Vision: 4th International Conference, SSVM 2013, Schloss Seggau, Leibnitz, Austria, June 2-6, 2013. Proceedings 4}, pages 380--391. Springer, 2013.

\bibitem{Xiao2018}
Chaowei Xiao, Ruizhi Deng, Bo~Li, Fisher Yu, Mingyan Liu, and Dawn Song.
\newblock Characterizing adversarial examples based on spatial consistency information for semantic segmentation.
\newblock In {\em European Conference on Computer Vision (ECCV)}, 2018.

\bibitem{xie2021segformer}
Enze Xie, Wenhai Wang, Zhiding Yu, Anima Anandkumar, Jose~M Alvarez, and Ping Luo.
\newblock Segformer: Simple and efficient design for semantic segmentation with transformers.
\newblock {\em Advances in Neural Information Processing Systems}, 34:12077--12090, 2021.

\bibitem{Xu2021}
Xiaogang Xu, Hengshuang Zhao, and Jiaya Jia.
\newblock Dynamic divide-and-conquer adversarial training for robust semantic segmentation.
\newblock In {\em IEEE International Conference on Computer Vision (ICCV)}, 2021.

\bibitem{b9}
Matthew~D Zeiler and Rob Fergus.
\newblock Visualizing and understanding convolutional networks.
\newblock In {\em Computer Vision--ECCV 2014: 13th European Conference, Zurich, Switzerland, September 6-12, 2014, Proceedings, Part I 13}, pages 818--833. Springer, 2014.

\bibitem{Zhang2021Delving}
Chongzhi Zhang, Mingyuan Zhang, Shanghang Zhang, Daisheng Jin, Qiang feng Zhou, Zhongang Cai, Haiyu Zhao, Shuai Yi, Xianglong Liu, and Ziwei Liu.
\newblock Delving deep into the generalization of vision transformers under distribution shifts.
\newblock {\em 2022 IEEE/CVF Conference on Computer Vision and Pattern Recognition (CVPR)}, pages 7267--7276, 2021.

\end{thebibliography}

\end{document}